\documentclass[10pt]{article}

\usepackage[preprint]{rlj} 

\usepackage{amssymb}            
\usepackage{mathtools}          
\usepackage{mathrsfs}           
\usepackage{graphicx}           
\usepackage{subcaption}         
\usepackage[space]{grffile}     
\usepackage{url}                
\usepackage{booktabs}
\usepackage{longtable}
\usepackage{multirow}
\usepackage{gensymb}
\usepackage{makecell}
\usepackage{wrapfig}
\usepackage{siunitx}
\usepackage{layouts}
\usepackage{csquotes}
\usepackage[table]{xcolor}

\newcommand{\abstracttext}[0]{%
Deep reinforcement learning has the potential to solve attitude control problems more adaptively, precisely, and robustly by handling nonlinear dynamics, uncertainties, and failure cases more effectively than traditional attitude control approaches.
We explore reinforcement learning (RL) for attitude control in spacecraft re-entry.
An industry-standard proportional-integral-derivative controller with gain scheduling serves as a strong baseline for model-free RL and hybrid controllers that combine these two approaches.
We formalize the application in the RL framework to apply continuous, off-policy RL.
State-of-the-art RL achieves comparable performance to traditional control approaches in this domain. However, its out-of-distribution generalization is not sufficient.
Hence, we use dynamics randomization to introduce challenging task variations during training and enforce generalization in a predefined operational envelope.
Finally, we assess the best obtained RL-based controllers with application-specific metrics to show superior performance in comparison to traditional controllers in the operational envelope, that is, hybrid controllers are able to track the angle of attack better and are more robust under variations of mass, inertia tensor, and flap actuator bandwidth.
}

\newcommand{\contributiontextone}[0]{%
We apply state-of-the-art deep reinforcement learning algorithms for continuous control to the problem of attitude control during hypersonic re-entry of a spacecraft. This is a challenging tracking problem that requires gain scheduling for changing atmospheric conditions.
We design a simple and effective reward function for attitude control and find that MR.Q excels without task-specific tuning.
}
\newcommand{\contributiontexttwo}[0]{%
Although the best learned controller performs better than the industry-standard baseline controller under nominal conditions, we explore hybrid control architectures that combine the baseline controller with reinforcement learning and compare these against pure reinforcement learning to enhance out-of-distribution generalization.
}
\newcommand{\contributiontextthree}[0]{%
To improve robustness and generalization explicitly, we compare dynamics randomization and task scheduling approaches for reinforcement learning in this attitude control problem and determine the best training strategy. We find that the resulting policies considerably improve robustness in comparison to the baseline controller.
}

\title{Deep Reinforcement Learning for Spacecraft Attitude Control During Atmospheric Re-Entry}

\setrunningtitle{Deep Reinforcement Learning for Attitude Control}

\author{Alexander Fabisch\textsuperscript{1}, Melvin Laux\textsuperscript{1}, Mariela De Lucas Álvarez\textsuperscript{1}, Edoardo Caroselli\textsuperscript{2}, Julian Theis\textsuperscript{2}}

\emails{\{alexander.fabisch,melvin.laux,mariela.de\_lucas\_alvarez\}@dfki.de, \{edoardo.caroselli,julian.theis\}@airbus.com}

\affiliations{
$^{1}$\textbf{Robotics Innovation Center, German Research Center for Artificial Intelligence (DFKI GmbH), Germany}\\
$^{2}$\textbf{Airbus Defence and Space GmbH, Germany}
}

\contribution{\contributiontextone}
{
We apply existing reinforcement learning algorithms (e.g., MR.Q \citep{fujimoto2025_mrq}). Prior work with older algorithms, less thorough evaluation, or more complicated reward functions in the application domain exists \citep[e.g.,][]{elkins_adaptive_2020}.
}

\contribution{\contributiontexttwo}
{
\citet{liu_attitude_2022} propose a similar hybrid controller that integrates reinforcement learning in a traditional controller. In comparison to this work, we use state-of-the-art algorithms and explicitly investigate generalization and robustness of learned controllers.
}

\contribution{\contributiontextthree}
{
Uniform dynamics randomization has been used successfully for various applications in robotics \citep{Antonova2017,Peng2018,Tan2018,OpenAI2020}.
Task scheduling approaches have been proposed in the context of deep reinforcement learning \citep{Cho2024} and contextual policy search \citep{Fabisch2014}.
}


\summary{\abstracttext}

\begin{document}

\makeCover
\maketitle

\begin{abstract}
\abstracttext
\end{abstract}

\section{Introduction}
\begin{wrapfigure}{r}{0.55\textwidth}
\centering
\vspace{-2.2cm}
\resizebox{0.55\textwidth}{!}{\input{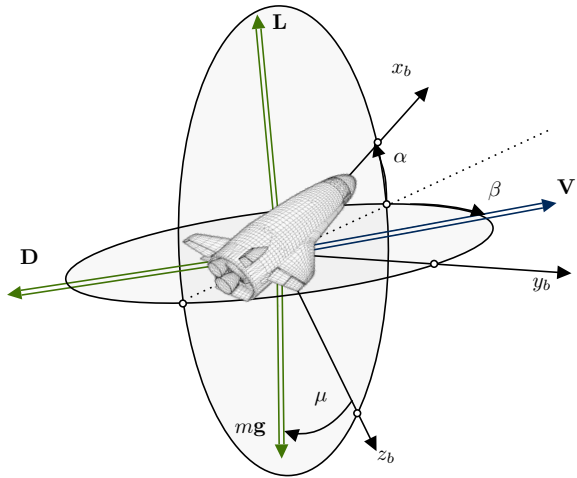}}
\caption{Sketch of the hypersonic re-entry vehicle.\label{fig:sketch}}
\vspace{-0.35cm}
\end{wrapfigure}

Deep reinforcement learning (RL) for continuous control \citep[e.g.,][]{lillicrap_continuous_2019} has the potential to handle nonlinear dynamics and failure cases more easily and to be more robust against and adaptable to unforeseen conditions than other attitude control approaches \citep[see, e.g.,][]{elkins_autonomous_2020,bernini_reinforcement_2024,liu_attitude_2022}.
However, the introduction of RL into a safety-critical attitude control system requires a fundamental shift in verification and validation (V\&V) for space-grade software. V\&V for guidance, navigation and control traditionally relies on deterministic and formally verifiable control laws, e.g., proportional–integral–derivative (PID) controllers. For these, the system's behavior is auditable through formal analysis and exhaustive simulation of well-defined corner cases.
Core challenges with deep RL are the out-of-distribution generalization and opacity of the learned policy.
It is a black box, making it infeasible to formally prove its stability and performance across the entire state space and parametric envelope. 
While explainable artificial intelligence advances transparency in the decision-making process of deep RL \citep{Li2025,Goel2025,Luss2023}, its integration into safety-critical control applications is not fully established.
We propose the use of hybrid control designs \citep[e.g., residual RL,][]{Johannink2019} to mitigate black-box concerns.

We specifically discuss attitude control for a hypersonic re-entry vehicle, as illustrated in Figure~\ref{fig:sketch}.
Such spacecraft return personnel and cargo from space and, hence, are a key technology for future human spaceflight, planetary exploration, and for establishing an extraterrestrial resource extraction economy, e.g., lunar mining.
We attempt on the one hand to replace classical control engineering by RL, reducing the expert's effort, and on the other hand to combine PID control with RL-based control to form a hybrid controller that more effectively handles challenging changes to the environment dynamics and failure cases.
We compare three different types of controllers.
(1)~Baseline controller: The previous solution is a PID controller with gain scheduling.
(2)~Only RL: RL without any additional controller, but with the model of the spacecraft to generate kinematically feasible reference values.
(3)~Hybrid controller: RL modifies the baseline controller.
The hybrid controller is our primary candidate for a flight-ready system. This architecture is inherently safer because the PID controller acts as a verified baseline. The RL agent's authority can be explicitly bounded by a supervisor logic that monitors the system's state and the RL agent's output, disengaging the RL agent when the vehicle approaches the edge of its validated safe domain.
We aim to provide statistical evidence that the controller is reliable and safe within a strictly defined operational envelope.

Our contributions are the following.
(1) \contributiontextone
(2) \contributiontexttwo
(3) \contributiontextthree

\section{Background: Spacecraft Re-Entry and Reinforcement Learning}
\label{sec:background}

\subsection{Spacecraft Re-Entry Problem}
\label{sec:environment}
The present paper considers flight path control for a lifting-body vehicle with low lift-to-drag ratio during hypersonic re-entry (see Figure~\ref{fig:sketch}).
We focus on the endo-atmospheric part of the flight trajectory and disregard the preceding deorbit maneuver.
Specifically, we focus on the descent from the upper mesosphere down to the troposphere (altitude \SI{93}{\kilo\meter} to \SI{10}{\kilo\meter}).
The vehicle speed at the start is \SI{7378}{\meter\per\second} and decreases to approximately \SI{150}{\meter\per\second} at the end, when a parachute can be deployed for landing.
Atmospheric flight creates aerodynamic forces and moments that act on the vehicle. The main objective is to control the aerodynamic forces such that they result in a desired translational motion of the re-entry vehicle.
In addition, the magnitude of the forces must remain limited to not endanger structural integrity of the vehicle.
These objectives are addressed by tracking a specific angle of attack and bank angle.
The angle of attack $\alpha$ describes the pitch attitude of the vehicle with respect to its direction of flight, while the bank angle $\mu$ describes the roll attitude of the vehicle with respect to this direction.  Furthermore, the nose of the vehicle needs to be aligned with its direction of flight, i.e., the sideslip angle $\beta$ must remain small.
Figure~\ref{fig:sketch} illustrates these angles.
Attitude control must minimize the errors between the commanded and actual angles over the course of the trajectory.
The vehicle in this study has two flaps at its tail that deflect to produce aerodynamic moments about the body-fixed axes $x_b$ and $y_b$, and thrusters arranged in a configuration to exert a moment about the axis $z_b$. The actions of attitude controllers therefore manipulate deflection commands to the aerodynamic control surfaces ($\delta_\text{e,cmd}$ for symmetric and $\delta_\text{a,cmd}$ for antisymmetric deflection) and a thruster command $\boldsymbol{\tau} = \left[\tau_x, \tau_y, \tau_z\right]^T$.
The observations encompass command values for angle of attack ($\alpha_\text{cmd}$) and bank angle ($\mu_\text{cmd}$) as well as estimates of the actual angles $\alpha,\beta,\mu$, and body-fixed angular rates from the navigation filter.

\subsection{Simulation Model}
\label{sec:simulation_model}
In order to train and evaluate the controllers, we rely on a simulation that encompasses the vehicle dynamics, actuator and sensor lags, atmosphere, as well as guidance, navigation, and control algorithms that serve as a baseline for this study. The models are briefly described next with additional details given in Appendix~\ref{app:sim_details}. 
The step size of the simulator is $1/140\,\mathrm{s}$ and it accounts for multi-rate subsystems and computational delay. For example, control algorithms are executed at \SI{14}{\hertz}.

A six-degree-of-freedom rigid-body flight dynamics model for a rotating, spherical earth is the core of the simulation \cite[see, e.g.,][]{stevens2003aircraft,stengel2022flight}.
In addition to the notation from Figure~\ref{fig:sketch}, the aerodynamic side force $Y$ (perpendicular to aerodynamic lift $L$ and drag $D$), and the vehicle mass $m$ are needed in the evolution of the flight path, defined in terms of velocity $V$, flight path angle $\gamma$ (vertical direction relative to the local horizontal), and course angle $\chi$ (horizontal direction relative to local north) as
\begin{equation}
\begin{bmatrix}
\dot{V} \\ \dot{\gamma}\\ \dot\chi   
\end{bmatrix}
= \frac{1}{m}
\begin{bmatrix}
    -D-mg\sin\gamma   
    \\ 
     \left(L\cos\mu - Y\sin\mu -mg\cos\gamma\right)/V
    \\ 
    \left( L\sin\mu + Y\cos\mu\right)/V 
\end{bmatrix}.
\label{eq:flightpath}
\end{equation}
The change of the aerodynamic angles $\alpha$, $\beta$, and $\mu$ is
\[
    \begin{bmatrix}
    \dot{\mu} \\ \dot{\alpha}\\ \dot\beta   
    \end{bmatrix}
    =
    \begin{bmatrix} (\omega_x \cos\alpha+\omega_z\sin\alpha)/\cos\beta - \dot{\chi}\sin\gamma+ \tan\beta (\dot{\gamma}\cos\mu +\dot{\chi}\cos\gamma\sin\mu) \\ \omega_y - \tan\beta (\omega_x\cos\alpha+\omega_z\sin\alpha) - (\dot{\gamma}\cos\mu +\dot{\chi}\cos\gamma\sin\mu)/(\cos\beta) \\ \omega_x\sin\alpha-\omega_z\cos\alpha + \dot{\chi}\cos\gamma\cos\mu-\dot{\gamma}\sin\mu
    \end{bmatrix}, 
\]
where we denote angular rates about the body-fixed axes by $\boldsymbol{\omega}=[\omega_x, \omega_y, \omega_z]^T$. The rotational dynamics are described by
$\boldsymbol{\dot{\omega}} =  \boldsymbol{I}^{-1} \left( \boldsymbol{M} - \boldsymbol{\omega} \times \boldsymbol{I} \, \boldsymbol{\omega} \right)$, where $\boldsymbol{M}\in\mathbb{R}^3$ denotes the moments resulting from aerodynamics and thruster usage and  $\boldsymbol{I}$ is the inertia tensor with respect to the center of mass.
The aerodynamic forces and moments $D,L,Y,\boldsymbol{M}$ are complicated to model accurately and are, in general, nonlinear functions of the vehicle state and environmental conditions. 
We employ an industry-grade high-fidelity aerodynamics model that represents these dependencies in terms of tabulated coefficients that depend on the flight conditions. This is a standard approach in flight dynamics modeling \citep[see, e.g.,][]{schmidt2011modern,stengel2022flight}. For example, the lift is modeled as 
$L = \bar{q} S C_L(M\!a,\alpha,\beta,\boldsymbol{\omega},\boldsymbol{\delta}).$
Here, $\bar{q}=\frac{1}{2} \rho V_a^2$ is the dynamic pressure, resulting from the velocity of the vehicle relative to air $V_a$ (i.e., $V$ accounted for wind) with the local air density $\rho$, and $S$ is the effective surface area of the vehicle. 
The coefficient $C_L$ depends on the Mach number $M\!a$ (the ratio of flight speed and local speed of sound) and the flap deflections $\boldsymbol{\delta}=[\delta_\text{e}, \delta_\text{a}]^T$ in addition to the state variables $\alpha$, $\beta$, and $\boldsymbol{\omega}$.
The simulation encompasses $M\!a\in [0.5, 26.8]$ and  $\bar{q}\in [50, 5825]\,\mathrm{Pa}$, leading to considerable variations in the aerodynamic influence.
The international standard atmosphere model is used to determine environmental conditions such as the density of air and speed of sound. 

Trajectory generation is a complex, mission-specific endeavor that requires precalculation and predictor-corrector algorithms \citep{lu_predictor_2008,vernis_accurate_2011,speng_robust_2011}. We consider a generic atmospheric re-entry mission with a simplified flight path guidance loop. One objective for the controller is to keep the sideslip angle small, $\beta\approx0$, which implies $Y\approx0$. For this case, Equation~\eqref{eq:flightpath} simplifies and provides insight into the principal mechanics behind guidance for re-entry.
The gravitational force ($mg$) accelerates the vehicle during descent ($\gamma<0$), while the aerodynamic drag $D$ decelerates the vehicle.
The change in $\gamma$ depends on the term $L\cos\mu$, which represents the upward component of the aerodynamic lift $L$ that opposes the gravitational force ($mg\cos\gamma$). Hence, the vertical direction of flight can be controlled by changing the magnitude of the bank angle $\mu$. However, the equation for $\chi$ reveals that this will also change the horizontal direction due to the sideward lift component $L\sin\mu$.
We employ a simple guidance strategy: 
(1)~The angle of attack determines the magnitude of lift and drag. The command value $\alpha_\text{cmd}$ issued by the guidance is a setpoint that ensures structural integrity of the vehicle, while resulting in sufficient deceleration along the trajectory.
(2)~The magnitude of the bank angle command $\mu_\text{cmd}$ is calculated by a feedback loop to track a predefined flight path angle $\gamma$, i.e., to offset the gravity component $mg\cos\gamma$ in Equation~\eqref{eq:flightpath}.
(3)~Bank angle reversal is triggered, i.e., the sign of the command value $\mu_\text{cmd}$ is switched, whenever a specified heading angle deviation from the reference is exceeded.
This strategy leads to sufficiently rich variations in the reference commands for the present study.

\subsection{Baseline Controller}
\label{sec:baseline}
We focus on the atmospheric control mode, which is more complex compared to exo-atmospheric attitude control. 
To provide a challenging baseline for benchmarking, a state-of-the-art control algorithm is included in the simulation. This baseline consist of separate control laws for the longitudinal motion ($\alpha\to\alpha_\text{cmd}$ tracking) and for the lateral-directional motion ($\mu\to\mu_\text{cmd}$ tracking and $\beta\to0$ regulation). Longitudinal control uses a gain-scheduled proportional-integral-derivative (PID) law in combination with an inverse model $f$ of the flight dynamics in feedforward:
\begin{eqnarray*}
    \delta_\text{e}(t)
    &=& f(M\!a(t),\bar{q}(t),\alpha_\text{cmd}(t)) +
    k_p(M\!a(t),\bar{q}(t))\cdot (\alpha_\text{cmd}(t)-\alpha(t))
    \\
    &&+\, k_i(M\!a(t),\bar{q}(t))\cdot \int^t_0 \alpha_\text{cmd}(s)-\alpha(s) ds
    - k_d(M\!a(t),\bar{q}(t))\cdot \dot\alpha(t)
\end{eqnarray*}
This is a deterministic, time-varying control policy with continuous changes of the controller gains based on the current estimates of the Mach number $M\!a(t)$ and dynamic pressure $\bar{q}(t)$.
Similarly, lateral-directional control laws are gain-scheduled PID for bank angle tracking and proportional-derivative (PD) control for sideslip angle regulation. The lateral directional control laws involve crossfeeds and hence they are multivariable. The controller aims at yielding as little as possible error between the commanded and the estimated values of the angle of attack, bank angle, and sideslip angle. The gains for the control laws were selected based on the well-established pole-placement technique that seeks to match a closed-loop response with desired characteristics specified by a low-order transfer function reference model \citep[see, e.g.,][]{enns1994dynamic,stevens2003aircraft}. Designs were performed for 21 linear snapshot models obtained along the considered nominal flight trajectory. The control laws are implemented in terms of state feedback gain schedules with linear interpolation in between design grid points. The scheduling variables are calculated based on estimated Mach number and dynamic pressure \citep[similar to][]{ganet_ARD_2008}.

\subsection{Reinforcement Learning for Continuous Control and Generalization of Policies}
Deep Deterministic Policy Gradients \citep[DDPG,][]{lillicrap_continuous_2019} enabled using neural networks for
continuous model-free RL.
Since then, many off-policy RL algorithms for continuous control were influenced by Soft Actor Critic \citep[SAC,][]{haarnoja_soft_2018} and DDPG's extension TD3 \citep{fujimoto18a_td3}. TD3 adds clipped double Q-learning, target policy smoothing, and delayed policy updates to DDPG. SAC is similar to TD3, but uses a stochastic policy in the maximum entropy RL framework.
TD3 was extended to TD7 \citep{Fujimoto2023_td7} and MR.Q \citep{fujimoto2025_mrq}. TD7 adds a state-action encoder, loss-adjusted prioritized experience replay \citep[LAP,][]{Fujimoto2020_lap}, and checkpointing. Similarly, MR.Q uses model-based representation learning and LAP.
SAC was extended to DroQ \citep{hiraoka2022_droq}, CrossQ \citep{Bhatt2024_crossq}, BRO \citep{Nauman2024_bro}, and SimbaV2 \citep{lee2025_simbav2}.
Across various benchmarks for continuous control, model-free RL \citep{Nauman2024_bro,fujimoto2025_mrq,lee2025_simbav2} currently seems to outperform model-based RL, i.e., DreamerV3 \citep{Hafner2025_dreamerv3} and TD-MPC2 \citep{hansen2024tdmpc}.

To obtain robust policies and enable sim-to-real transfer, dynamics randomization has been used in robotics \citep{Antonova2017,Peng2018,Tan2018,OpenAI2020}. In its simplest form, dynamics randomization varies parameters of the dynamics in the simulation randomly during training without informing the RL agent. The agent has to learn a policy that is robust under these unknown variations while only relying on observations.

We have a smoothly varying parametrization of the dynamic properties over which we want to generalize, e.g., actuation characteristics.
Hence, we adopt the formalism of Contextual Markov Decision Processes \citep[CMDPs,][]{Hallak2015Contextual}, in which the variation of tasks is described as a tuple
$\left(\mathcal{C}, \mathcal{S}, \mathcal{A}, \mathcal{M}(c)\right),$
with context space $\mathcal{C}$, state space $\mathcal{S}$, action space $\mathcal{A}$, and a function $\mathcal{M}(c) = \left( \mathcal{S}, \mathcal{A}, P_c(s'|s, a), R_c(r|s, a, s'), \mu_c(s_0) \right)$ mapping contexts $c \in \mathcal{C}$ to individual Markov Decision Processes (MDPs)
with context-dependent state transition probabilities $P_c$, reward function $R_c$, and initial state distribution $\mu_c$. In our case, the agent only has access to observations that depend on context and previous states $o_t = \phi(c, s_t, s_{t-1}, \ldots)$, and it is supposed to select actions $a \in \mathcal{A}$.
Our goal is to find a policy \(\pi\) that maximizes the objective function 
$\frac{1}{|\mathcal{C}|} \sum_{c \in \mathcal{C}} J(\mathcal{M}(c), \pi),$
where \(J(\mathcal{M}, \pi)\) is the expected return of policy \(\pi\) in task \(\mathcal{M}\).
CMDPs allow us to define distinct sets of contexts for training and testing to evaluate zero-shot generalization \citep{Kirk2023}.

Going one step further, we want to select task variations intelligently during training to improve sample efficiency and final performance. Task scheduling approaches have been proposed for multi-task RL \citep{Sharma2018AMT,Cho2024} and contextual RL \citep{Fabisch2014}. \citet{Fabisch2014} frame the task selection problem as a non-stationary multi-armed bandit problem and solve it via discounted upper-confidence bound \citep[D-UCB,][]{Kocsis2006,Garivier2011}. In this framework, they compare task selection strategies based on the episodic return: easiest tasks first (best reward), hardest task first, and focus on tasks with largest improvement in returns (e.g., monotonic progress). We call this active multi-task training (AMT). Scheduled multi-task training \citep[SMT,][]{Cho2024} focuses on solving hard tasks first.

\section{Learning Attitude Control for Hypersonic Spacecraft Re-Entry}
\label{sec:applying_rl}

\paragraph{Action space (see also Appendix \ref{sec:app_actions})}
The vehicle uses its flaps to control pitch and roll motion, and thrusters for yaw motion. We compare several architectures for attitude control. The simplest learned controller (Only RL) will directly output yaw thrust and changes of the flap angles.
The controller described in Section~\ref{sec:baseline} serves as a baseline for evaluation, but also as a building block in two hybrid controllers.
In additive hybrid control, the output of the RL policy network is added to the output of the baseline controller \citep[also known as residual RL,][]{Johannink2019}.
A more advanced hybrid control approach uses the policy for gain scheduling in the baseline controller, i.e., we compute twelve continuous gain factors $f_\text{gain}$ to scale gains in the baseline control laws.
We represent these multipliers as values of decibel in accordance with classical control notion, i.e., we transform each policy output by $f_\text{gain}(x) = 10^{x/20}$ with $x \in \left[-6, 6\right]$, such that a policy output of $x=0$ results in a factor of 1. The maximum is $f_\text{gain}(6)\approx2$, and the minimum is $f_\text{gain}(-6)\approx 0.5$.

\paragraph{Observation space (see also Appendix \ref{sec:app_observations})}
The observable state of the vehicle includes the aerodynamic angles ($\alpha, \beta, \mu$), and angular velocities $\boldsymbol{\omega}$ (see Section \ref{sec:background}).
In addition, we add the commanded aerodynamic angles ($\alpha_{\text{cmd},t}, \beta_{\text{cmd},t}, \mu_{\text{cmd},t}$), the previously commanded flap deflections ($\delta_{e,\text{cmd},t-1},\delta_{a,\text{cmd},t-1}$), and, for hybrid control, also the current baseline control command to the observation space.
To mimic the structure of a conventional PID error-feedback controller, we add for each aerodynamic angle the current error, its derivative, and its integral to the observation space. This improves the ability of a feedforward policy network in this partially observable MDP.
The dynamics of the environment are influenced by air density and speed of the vehicle, which determine the dynamic pressure. We measure the altitude, velocity, and dynamic pressure and provide it as observations to the controller.
Observations are normalized to the range $\left[-1, 1\right]$.

\paragraph{Reward function (see also Appendix \ref{sec:app_reward})}
Our main objective is to minimize the attitude cost
\[
c(t)
=
\left(\frac{\alpha_\mathrm{cmd}(t) - \alpha(t)}{\alpha_{\text{range}}}\right)^2
+
\left(\frac{\beta_\mathrm{cmd}(t) - \beta(t)}{\beta_{\text{range}}}\right)^2
+
\left(\frac{\mu_\mathrm{cmd}(t) - \mu(t)}{\mu_{\text{range}}}\right)^2,
\]
which we define based on commands and estimates of the aerodynamic angles (in radian).
The individual ranges allow us to define a desired range for each error. This approach is known as \emph{Bryson's rule} in classical linear quadratic optimal control. We focus the attitude cost on the angle of attack $\alpha$ with $\alpha_{\text{range}} = 2^\circ, \beta_{\text{range}}=10^\circ,\mu_{\text{range}}=10^\circ$.

However, taking the negative attitude cost as a reward has the drawback that the reward is always negative, which makes early termination of an episode by running into failure states appealing.
We solve this problem by defining a strictly positive reward
\[
\text{attitude reward}\, (t) = \max (0.001, \exp(-w_c \cdot c(t))) \in \left[0.001, 1\right]
\]
with weight $w_c=2$. Since the attitude reward is at least 0.001, it discourages the agent from terminating early and shifts focus to the main objective.

We combine the attitude reward with a control cost to form the reward function in each step $t'$ as
$r_{t'} = \text{attitude reward}\, (t) - \text{control cost}\, (t),$
where control cost is the sum of thruster and flap costs
\[
\text{control cost}\,(t) =
w_{\tau_z} \left(\frac{\tau_z(t)}{\tau_{z,\max}}\right)^2
+ w_{\delta_e} \left(\frac{\Delta\delta_{e,\text{cmd}}(t)}{\Delta \delta_{e,\max}}\right)^2
+ w_{\delta_a} \left(\frac{\Delta\delta_{a,\text{cmd}}(t)}{\Delta \delta_{a,\max}}\right)^2,
\]
with weights $w_{\tau_z} = 1$ and $w_{\delta_e} = w_{\delta_a} = 0.05$. 
The difference between successive flap commands $\Delta \delta_{\circ,\text{cmd}}(t) = \delta_{\circ,\text{cmd}}(t) - \delta_{\circ,\text{cmd}}(t-1)$ has a maximum of $\Delta \delta_{\circ,\max} = \Delta t \cdot \dot{\delta}_{\circ,\max} = \frac{1}{14}\, \cdot$ \SI{15}{\degree} $\approx$ \SI{1.07}{\degree}. The yaw torque $\tau_z(t)$ generated by the thrusters is limited by $\tau_{z,\max} = $ \SI{300}{\newton}.

\paragraph{Context space (see also Appendix \ref{app:sim_details})}
For dynamics randomization, we parametrize the simulation with the mass of the vehicle $m_0 \in \left[1312, 1968\right]$ \si{\kilo\gram}, fractions of the principal moments of inertia $\boldsymbol{f} \in \left[0.9, 1.1\right]^3$, misalignment of the principal axis by means of rotation $\boldsymbol{\omega} \in \{\omega = \theta\hat{\omega} | \hat{\omega} \in S^2, \theta \in \left[-10^\circ, 10^\circ\right] \} \subset \mathbb{R}^3$, and flap actuator bandwidth $\omega_0 \in \left[12, 30 \right]$ \si{\radian\per\second}. Hence, the parameter space is $\mathcal{C} \subset \mathbb{R}^8$. Nominal conditions correspond to $c_\text{nominal} = \left(1640, 1, 1, 1, 0, 0, 0, 30\right)^T$.

\section{Experiments}
\label{sec:experiments}

In our experiments, we want to answer the following research questions:
\begin{enumerate}[noitemsep,topsep=0pt]
\item Can we apply deep RL algorithms to spacecraft attitude control? (Section \ref{sec:rl_alg_selection})
\item What is the best achievable performance with pure RL? (Section \ref{sec:rl_alg_selection})
\item Which control architecture works best? (Section \ref{sec:ctrl_arch})
\item How well do controllers generalize to out-of-distribution conditions? (Section \ref{sec:ood_generalization_st})
\item Can we enforce generalization with domain randomization or task scheduling? (Section~\ref{sec:mtt})
\item Which is the most robust controller? (Section \ref{sec:mt_robust})
\end{enumerate}

\paragraph{Evaluation of RL algorithms:}
Following \citet{Agarwal2021_statistical_precipe}, we analyze the performance of RL algorithms with sample-efficiency curves using the interquartile mean (IQM) and bootstrapped 95\% confidence intervals of the cumulative episode reward, i.e., undiscounted return. For direct comparison of policies, we compute the probability of improvement.

\paragraph{Application-specific performance metrics:}
Key metrics for attitude control are the absolute errors
of the angle of attack $|e_\alpha(t)| = |\alpha_{\text{cmd}}(t) - \alpha(t)|$,
sideslip angle $|e_\beta(t)| = |\beta_{\text{cmd}}(t) - \beta(t)|$,
and bank angle $|e_\mu(t)| = |\mu_{\text{cmd}}(t) - \mu(t)|$.
We prioritize the angle of attack to maintain structural and thermal integrity of the vehicle and create sufficient drag to decelerate. We want to avoid large errors, while small errors over a longer period of time are allowed. Hence, we analyze the error distribution and report percentiles.
As secondary metrics, we report the control costs in all three action components.
For multiple test contexts, we furthermore report the success rate, which is the percentage of contexts, in which the controller was able to follow the trajectory completely.
Episodes terminate when any angle is outside of its safe domain, which is defined based on the aerodynamic angles: $\alpha \in \left[0^{\circ}, 60^{\circ}\right]$, $\beta \in \left[-20^{\circ}, 20^{\circ}\right]$, and $\mu \in \left[-90^{\circ}, 90^{\circ}\right]$.

\subsection{Training Under Nominal Conditions}

We start episodes with constant initial attitude, initial mass, inertia, and flap actuator bandwidth, but randomize parameters of the trajectory to prevent overfitting the trajectory (see Appendix \ref{app:sim_details} for details).
We analyze which control architecture and RL algorithm performs best and what is the maximum achievable performance. Since we do not vary dynamics during training, we can test out-of-distribution generalization of the learned controllers.

\subsubsection{RL Algorithm Selection}
\label{sec:rl_alg_selection}
\begin{figure}[t]
\centering
\includegraphics{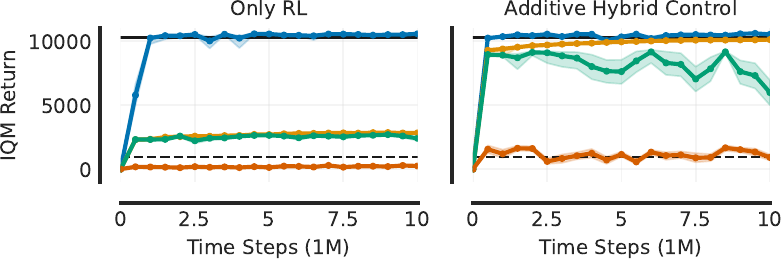}\\
\includegraphics{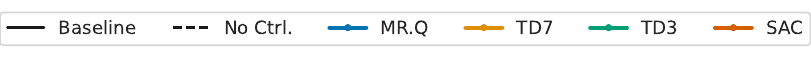}
\vspace{-0.4cm}
\caption{MR.Q surpasses the baseline controller. Learning curves with interquartile means (IQM) and 95\% bootstrap confidence intervals for 10 training seeds \citep{Agarwal2021_statistical_precipe}.
\label{fig:learning_curve}
}
\vspace{-0.6cm}
\end{figure}

We initially selected SAC and TD3 to explore the feasibility RL-based control. We tried CrossQ as an extension of SAC as well as TD7 and MR.Q as extensions of TD3. In this study, we focus on TD3-based algorithms as they showed consistent behavior in preliminary experiments without task-specific tuning. Hence, we evaluate only SAC, TD3, TD7, MR.Q.
Hyperparameters are listed in Appendix \ref{sec:app_hyperparams}.

\begin{wrapfigure}{r}{0.5\textwidth}
\centering
\vskip -0.9cm
\begin{subfigure}[c]{0.5\textwidth}
\includegraphics[width=\textwidth]{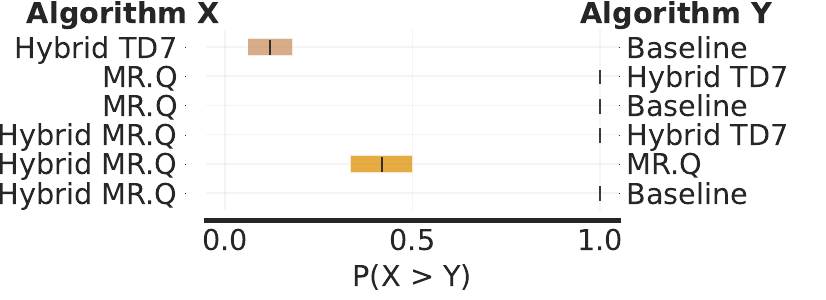}
\vskip -0.2cm
\subcaption{Probability of improvement \citep{Agarwal2021_statistical_precipe}.\label{fig:st_pi}}
\end{subfigure}
\vskip 0.2cm
\begin{subfigure}[c]{0.5\textwidth}
\includegraphics[width=\textwidth]{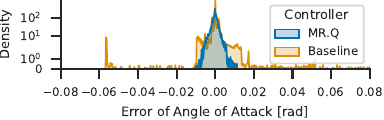}
\vskip -0.1cm
\subcaption{Error distribution (kernel density estimation) for angle of attack. Density is plotted on a logarithmic scale to highlight the tails. The peak of the baseline controller around $0$ is more pronounced on a linear scale.\label{fig:st_error_dist}}
\end{subfigure}
\vspace{-0.2cm}
\caption{Comparison under nominal conditions.}
\vspace{-0.6cm}
\end{wrapfigure}

With the learning curves in Figure \ref{fig:learning_curve}, we evaluate the performance and sample efficiency of RL algorithms and control architectures during training.
We evaluate combinations of learning algorithm and control architecture with 10 random seeds for training and 10 evaluation episodes without exploration noise.
For comparison, we plot the performance of no control and the baseline controller averaged over 10 evaluation episodes. Individual learning curves per training run are in Appendix \ref{app:individual_learning_curves}.

MR.Q excels without task-specific tuning.
The main reasons for this are the multi-step return, which reduces bias in the target value for the critic loss, and the model-based representation (detailed analysis in Appendix~\ref{app:ablations}).
Most algorithms perform better in the hybrid control architecture than their non-hybrid counterpart, with the exception of MR.Q.
For the non-obvious cases, we compare the best obtained policies per run with the probability of improvement and a 95\% confidence interval in Figure~\ref{fig:st_pi}.
We attribute that MR.Q outperforms the baseline controller to the fact that MR.Q learns to continuously adapt to environment conditions, which is more efficient than the gain scheduling of the baseline.
In particular, the error distribution for our primary control objective, the angle of attack, is narrower for the learned controller than for the baseline (see Figure~\ref{fig:st_error_dist}, see also Appendix~\ref{app:performance_single_task} for obtained trajectories and control commands). However, the baseline has a higher concentration of errors close to 0.
Hence, MR.Q works without task-specific tuning and is able to achieve a higher return than the baseline controller by tracking the angle of attack more accurately.

\subsubsection{Comparison of Control Architectures}
\label{sec:ctrl_arch}
\begin{wrapfigure}{r}{0.48\textwidth}
\centering
\vspace{-1cm}
\includegraphics{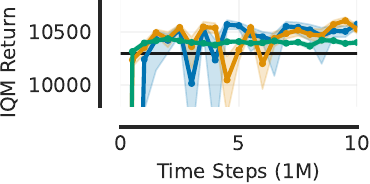}\\[0.05cm]
\includegraphics[width=0.48\textwidth]{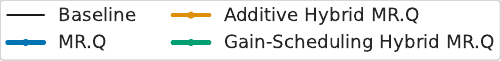}
\vskip -0.2cm
\caption{Comparisons of control architectures.\label{fig:architecture_comparison}}
\vspace{-0.5cm}
\end{wrapfigure}
With MR.Q, we compare control architectures (see Figure \ref{fig:architecture_comparison}).
Our simplest RL-based controller maps from observations to flap changes and thruster commands. We compare it to additive and a gain-scheduling hybrid control.
In both hybrid architectures, the baseline controller's command is in the observation space.

For maximum performance pure RL or additive hybrid control seem to be the best options under nominal evaluation conditions.
Although the best possible performance of the gain-scheduling hybrid controller is lower than for the two other architectures, the bounded influence of the policy makes learning considerably more stable across runs and over time, and the performance is still better than the baseline.

\subsubsection{Out-Of-Distribution Generalization of Policies Trained Under Nominal Conditions}
\label{sec:ood_generalization_st}
\begin{figure*}[t!]
\centering
\begin{subfigure}[c]{\textwidth}
\centering
\includegraphics[width=\textwidth]{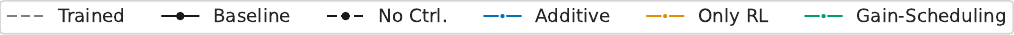}
\rotatebox{90}{\hspace{2.5cm}Return (Undiscounted)}
\includegraphics{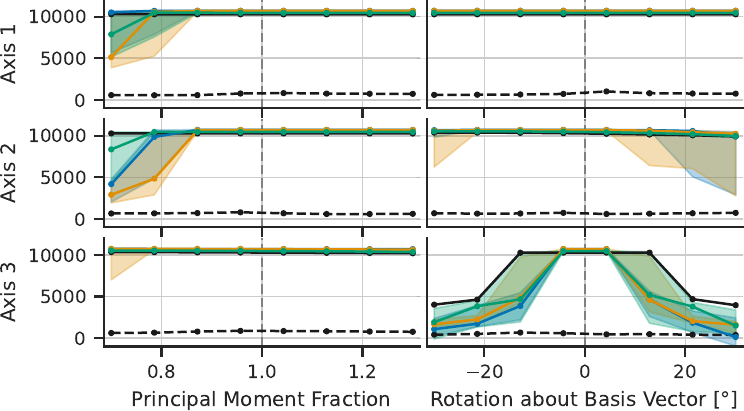}
\vspace{-0.1cm}
\subcaption{Performance evaluated under variations of \(\boldsymbol{I}\). On the left, we scale the principal moments of inertia along three axes (from top to bottom: x-, y-, z-axis). On the right, we rotate \(\boldsymbol{I}\) about the three basis vectors.}
\label{fig:st_inertia}
\end{subfigure}

\vskip 0.1cm

\begin{subfigure}[c]{\textwidth}
\centering
\includegraphics{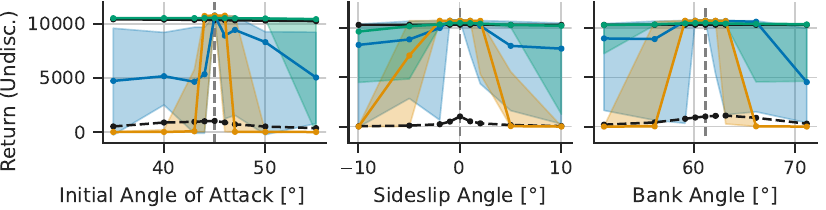}
\vspace{-0.6cm}
\subcaption{Performance with respect to changes of the initial attitude without training for it. Performance is measured at the initial attitude during training $+\, \epsilon \in \left[-10, -5, -2, -1, 0, 1, 2, 5, 10\right]$ degree per component.}
\label{fig:figure_attitude_variations}
\end{subfigure}

\vskip 0.1cm

\begin{subfigure}[c]{\textwidth}
\centering
\includegraphics{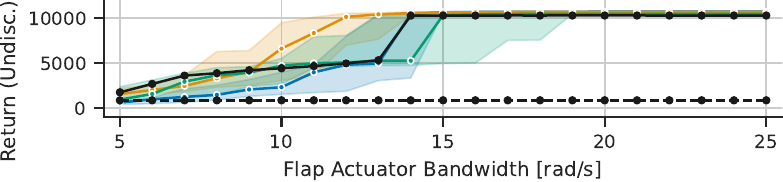}
\vspace{-0.1cm}
\subcaption{Performance evaluated under variations of flap actuator bandwidth.}
\label{fig:st_flap_frequency}
\end{subfigure}
\vspace{-0.2cm}
\caption{Generalization of policies trained under nominal conditions.
The solid lines show median performance over training runs and shaded areas the $\left[5, 95\right]$-percentile interval.}\label{fig:ood}
\vspace{-0.3cm}
\end{figure*}

To analyze generalization of the standard RL training paradigm with random exploration in action space, we test robustness with respect to variations that were not present during training.
We compare the undiscounted return of an episode for the best obtained policies per run (see Figure \ref{fig:ood}).

\textbf{Inertia tensor (Figure \ref{fig:st_inertia}):}
The inertia tensor \(\boldsymbol{I}\) has six degrees of freedom. To systematically vary it, we multiply the principal moments of inertia $I_{xx},I_{yy},I_{zz}$ by factors from $\left[0.7,1.3\right]$ and apply rotations about the basis vectors with angles from $\left[-30^\circ, 30^\circ\right]$. Rotation of \(\boldsymbol{I}\) about the z-axis has the most impact and the baseline controller is more robust than the learned controllers to it.

\textbf{Initial attitude (Figure \ref{fig:figure_attitude_variations}):}
We vary the initial values of the aerodynamic angles.
Although policies trained on constant initial attitude outperform the baseline under nominal conditions, they do not generalize well. Even for a small envelope around nominal conditions, we cannot guarantee stability without explicit testing, which demonstrates that action noise is not enough to learn policies that are robust against deviations from nominal conditions.
However, the additive hybrid controller's performance deteriorates less drastically in most out-of-distribution cases than pure RL and the gain-scheduling hybrid controller is almost as reliable as the baseline.

\textbf{Flap actuator bandwidth (Figure \ref{fig:st_flap_frequency}):}
Similarly, we test generalization to variations of the flap actuator bandwidth. Lower values make the control problem harder and simulate failures of the actuation. Nominal conditions correspond to \SI{30}{\radian\per\second}. Hence, all of the tested values are out of distribution. The baseline controller's performance drastically drops for actuator bandwidths below \SI{14}{\radian\per\second}. Interestingly, completely learned controllers are consistently more robust to lower values. We suspect that random exploration in action space during training makes the resulting policy robust to this scenario although the exact stable range cannot be guaranteed.
Both hybrid control architectures show more variation, but their median performance is similar to the baseline.

The bounded influence of the gain-scheduling hybrid controller makes its behavior more similar to the baseline with respect to variations of the initial conditions.
However, the stability of learned controllers cannot be guaranteed in the same range as for the baseline controllers when we do not explicitly enforce the operational envelope during training.
Although MR.Q is more robust to actuation failures than the baseline controller, it is less stable with respect to variations of the inertia tensor and considerably less stable under variations of the initial attitude than the baseline controller.

\subsection{Training Under Dynamics Randomization}
We randomize not just the trajectories, but also the initial attitude to ensure that the RL-based controllers generalize to these variations. Furthermore, we systematically vary the following dynamics parameters: initial mass, inertia tensor, and flap actuator bandwidth (see Appendix \ref{app:sim_details} for details).
Each combination of dynamics parameters defines a context. We compare task scheduling approaches to select contexts during training and evaluate generalization of the obtained controllers.

To test generalization and robustness of controllers, we uniformly sample a test set $\mathcal{C}_{\text{text}}$ of 100 contexts and measure the obtained return in each test context.
We select the best controllers per training procedure for this analysis based on the minimum return in a smaller test set of 30 contexts from regularly saved checkpoints per run (checkpoint interval: 200k time steps).

\subsubsection{Comparison of Task Scheduling Methods for Dynamics Randomization}
\label{sec:mtt}
We sample $|\mathcal{C}_\text{train}|=50$ context vectors that define individual tasks. We compare SMT, which prefers hard tasks, and active task selection with the
best-reward (AMT-B, prefers easy tasks) and monotonic progress (AMT-M, prefers tasks with high expected learning progress) strategies.
Hyperparameters are listed in Appendix \ref{sec:app_ts_hyperparams}.
We use continuous uniform random sampling of contexts (dynamics randomization, DR) and round robin (RR) as baselines for task scheduling.

\begin{wrapfigure}{r}{0.45\textwidth}
\centering
\vskip -0.5cm
\includegraphics[width=0.4\textwidth]{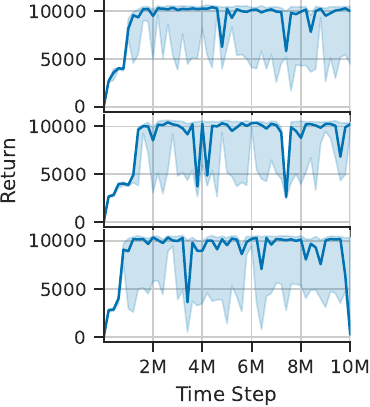}
\vskip -0.2cm
\caption{Learning curves of three training runs with DR and pure RL with MR.Q. Lines indicate median performance over 30 test contexts and shaded areas show the full range of returns.\label{fig:mt_learning_curve}}
\vskip -0.5cm
\end{wrapfigure}

Task scheduling methods do not perform better than dynamics randomization or round robin.
Since the results are inconclusive, we report them in Appendix~\ref{app:mt_metrics}.
Among task selection strategies, focusing on hard tasks in the beginning (SMT) is the worst strategy. Focusing on the easiest tasks first (AMT-B) and focusing on tasks with the maximum expected learning progress (AMT-M) work equally well and are comparable to the baselines DR and RR with no clear advantage.

Training with task scheduling is difficult for two reasons: (1) training on the full range of flap actuator bandwidths creates some problems that are considerably harder to solve, and (2) as we use a replay buffer for each task, infrequently selected tasks might contain transitions of low performing policies that are still selected relatively often, since we first sample the task uniformly and then the samples within the task for the batch that is required in the network updates. We mitigate the first issue by sampling flap actuator bandwidths from $\left[25, 30\right]$ \si{\radian\per\second}, but still evaluate on the full range. The second problem causes instabilities after about 4M steps, which we avoid by stopping after 5M steps. A more detailed analysis is required to understand the problem and find solutions.

\subsubsection{Generalization and Robustness of Best Controllers}
\label{sec:mt_robust}
\begin{table}[b!]
\vskip -0.4cm
\centering
\caption{Evaluation of best controllers with domain-specific metrics. Cell background color indicates \colorbox{green!40}{best values} or \colorbox{green!15}{values better than the baseline}. Appendix~\ref{app:mt_percentiles} contains more details on error distributions and Appendix~\ref{app:mt_metrics} contains all results with control costs.\label{tab:robustness}}
\vspace{-0.65cm}
{\scriptsize
\begin{longtable}{llr|llll|llll|llll}
\toprule
Training & \rotatebox{90}{Seed} & \rotatebox{90}{Success} & \rotatebox{90}{Median} & \rotatebox{90}{90-Perc.} & \rotatebox{90}{95-Perc.} & \rotatebox{90}{98-Perc.} & \rotatebox{90}{Median} & \rotatebox{90}{90-Perc.} & \rotatebox{90}{95-Perc.} & \rotatebox{90}{98-Perc.} & \rotatebox{90}{Median} & \rotatebox{90}{90-Perc.} & \rotatebox{90}{95-Perc.} & \rotatebox{90}{98-Perc.} \\
\midrule
& & &\multicolumn{4}{c}{$|e_\alpha|$ [deg]}&\multicolumn{4}{c}{$|e_\beta|$ [deg]} &\multicolumn{4}{c}{$|e_\mu|$ [deg]}\\
\midrule
Baseline & - & 79\% & \cellcolor{green!40}0.05 & 0.46 & 0.67 & 2.96 & \cellcolor{green!40}0.02 & \cellcolor{green!40}0.20 & \cellcolor{green!40}0.36 & \cellcolor{green!40}0.57 & \cellcolor{green!40}0.06 & 0.72 & 1.43 & 2.35 \\
\midrule
Add. Hyb. & 0 & \cellcolor{green!15}93\% & \cellcolor{green!40}0.05 & \cellcolor{green!40}0.23 & \cellcolor{green!40}0.34 & \cellcolor{green!40}0.59 & 0.07 & 0.25 & 0.41 & 0.65 & 0.17 & \cellcolor{green!40}0.42 & \cellcolor{green!40}0.59 & \cellcolor{green!40}0.87 \\
Add. Hyb. & 1 & \cellcolor{green!15}93\% & 0.07 & \cellcolor{green!15}0.37 & \cellcolor{green!15}0.56 & \cellcolor{green!15}0.83 & 0.07 & 0.30 & 0.53 & 0.85 & 0.15 & \cellcolor{green!15}0.60 & \cellcolor{green!15}0.84 & \cellcolor{green!15}1.11 \\
Add. Hyb. & 2 & \cellcolor{green!15}94\% & 0.07 & \cellcolor{green!15}0.33 & \cellcolor{green!15}0.47 & \cellcolor{green!15}0.71 & 0.11 & 0.39 & 0.60 & 0.84 & 0.17 & \cellcolor{green!15}0.60 & \cellcolor{green!15}0.84 & \cellcolor{green!15}1.19 \\
\midrule
Only RL & 0 & \cellcolor{green!40}100\% & 0.12 & 0.47 & \cellcolor{green!15}0.63 & \cellcolor{green!15}0.86 & 0.09 & 0.45 & 0.88 & 1.87 & 0.23 & 1.14 & 1.63 & \cellcolor{green!15}2.19 \\
Only RL & 1 & \cellcolor{green!15}99\% & 0.10 & \cellcolor{green!15}0.45 & \cellcolor{green!15}0.62 & \cellcolor{green!15}0.84 & 0.10 & 0.53 & 1.06 & 2.03 & 0.34 & 1.31 & 1.82 & 2.57 \\
Only RL & 2 & \cellcolor{green!15}98\% & 0.10 & \cellcolor{green!15}0.39 & \cellcolor{green!15}0.55 & \cellcolor{green!15}0.80 & 0.09 & 0.49 & 0.95 & 1.76 & 0.25 & \cellcolor{green!15}0.69 & \cellcolor{green!15}0.98 & \cellcolor{green!15}1.49 \\
\bottomrule
\end{longtable}
}
\end{table}

Since task scheduling approaches do not provide any benefit in comparison to continuous uniform dynamics randomization (DR), we analyze more closely the performance of the best obtained controllers with DR. We compare DR for pure RL and additive hybrid control in Table~\ref{tab:robustness} with domain-specific evaluation metrics. We focus on the error distribution of the aerodynamic angles and success rate. See Appendix~\ref{app:mt_metrics} for a detailed analysis including the control costs.
Exemplary training curves are displayed in Figure~\ref{fig:mt_learning_curve}).

In comparison to the baseline controller, which fails particularly often at low flap actuator bandwidths and large rotations of the inertia tensor, both additive hybrid control and pure RL improve the success rate considerably.
The additive hybrid controller's success rate seems to be limited by the baseline, although it obtains lower error 9x-percentiles in each aerodynamic angle.
For the angle of attack, both architectures reduce the 90, 95, and 98 error percentiles in comparison to the baseline.
The additive hybrid controller is even better at tracking the bank angle than the baseline in these percentiles.
Improving these metrics comes at the cost of more expensive control, i.e., thruster usage and flap angle changes (see Appendix~\ref{app:mt_metrics}).

\section{Related Work}
\label{sec:related_work}
RL has been applied to attitude control in various ways for vehicles inside \citep{zhen_deep_2020,bernini_few_2021,liu_attitude_2022,rosa_deep_2023,bernini_reinforcement_2024,bohn_data-efficient_2024,AI4GNC_City2024} and outside of an atmosphere \citep{su_deep_2019,vedant_reinforcement_2019,elkins_autonomous_2020,elkins_adaptive_2020,elkins_bridging_2021,mahfouz_reinforcement_2022,liu_neural_2022,xiao_fixed-time_2023,Barrenechea2023Hoppa}.
Q-learning \citep{liu_neural_2022}, PPO \citep{vedant_reinforcement_2019,elkins_autonomous_2020,zhen_deep_2020,mahfouz_reinforcement_2022}, DDPG \citep{su_deep_2019,rosa_deep_2023}, TD3 \citep{elkins_adaptive_2020,elkins_bridging_2021,liu_attitude_2022}, and SAC \citep{bernini_reinforcement_2024,bohn_data-efficient_2024} were used in these applications.
Most studies replace traditional control approaches by RL.
However, \citet{Barrenechea2023Hoppa} propose a residual RL architecture for online adaptation and
\citet{liu_attitude_2022} embed a policy for anti-disturbance control in a hybrid controller for hypersonic re-entry. Furthermore, \citet{liu_attitude_2022} extend TD3 to EVE-TD3, which shares similarities with MR.Q:
(1)~a multi-step return (over a horizon of 30 steps) to improve the estimation of the target value for the critic,
and
(2)~the policy network has a bottleneck (6--18 nodes) to force the policy to extract high-level features.

According to \citet{elkins_adaptive_2020}, robustness of attitude control can be defined with respect to
(1)~tumble: nonzero initial angular velocity,
(2)~single, impulsive disturbance torque,
(3)~constant disturbance torque,
and (4)~different inertia tensors $\boldsymbol{I}$.
RL can be robust against variations of $\boldsymbol{I}$ by adapting to it online \citep[e.g., in takeover maneuvers,][]{liu_neural_2022} or by handling variation of it as noise \citep{vedant_reinforcement_2019}.
Robustness can also be defined in terms of handling disturbances caused by the applications of unknown external torques on the spacecraft \citep{xiao_fixed-time_2023,elkins_adaptive_2020} or actuator faults \citep{xiao_fixed-time_2023}. Sources of disturbances might be
(1) fault caused by bias (e.g., less available torque),
(2) fault caused by failure to respond to control signals (e.g., motor fault), or
(3) disturbance of environment: uncertainties in air resistance, gravity, solar pressure, etc. \citep{su_deep_2019}.
\citet{bernini_reinforcement_2024} measure robustness with respect to wind gusts in quadcopter control, which is similar to a single, impulsive disturbance force.
Similarly, \citet{rosa_deep_2023} verified a deep RL algorithm for guidance and control of a reusable launch vehicle in the landing phase and tested under various wind conditions.

\section{Conclusion}
\label{sec:discussion}

The MR.Q algorithm excels without task-specific tuning in the challenging attitude control problem of a hypersonic re-entry vehicle (see Section~\ref{sec:rl_alg_selection}). Specifically, its model-based representation and multi-step returns are pivotal.
Pure MR.Q can surpass the performance of the baseline controller and obtain a similar performance as hybrid controllers under nominal conditions.
However, training under nominal conditions does not yield sufficient generalization, although hybrid control architectures often work better under out-of-distribution settings than pure RL (see Section~\ref{sec:ood_generalization_st}).
Hence, we enforce generalization with dynamics randomization and find that continuous uniform sampling of dynamic parameters is sufficient, i.e., task scheduling approaches do not perform better (see Section~\ref{sec:mtt}).
Surprisingly, MR.Q is able to generalize over each condition that we varied during training. With dynamics randomization, obtained controllers are considerably more robust than the baseline with respect to uncertainties in dynamics parameters (see Section~\ref{sec:mt_robust}).

The additive hybrid controller is the most promising controller. It generalizes better than pure MR.Q in the out-of-distribution experiments with training under nominal conditions (see Section~\ref{sec:ood_generalization_st}). It improves the robustness in comparison to the baseline by reducing the probability of larger errors of the angle of attack at the cost of using the thrusters and changing the flap angles more (see Section~\ref{sec:mt_robust}). The lower bound of the additive hybrid controller's performance is the baseline controller's performance if we deactivate the learned component in out-of-distribution settings.

To ensure stability of hybrid controllers, out-of-distribution conditions must be detected to disable the learned component and fall back to the formally verified baseline. This is straightforward to implement through outlier detection with data from the replay buffer.
To enable real-time execution of policies in deployment, we plan to run them on Field Programmable Gate Arrays (FPGAs).

Being able to adapt to unforeseen conditions online would be a major advantage of reinforcement learning for attitude control \citep[see, e.g.,][]{Barrenechea2023Hoppa}.
The gain-scheduling controller, which exhibits less variance across training steps and runs (see Section~\ref{sec:ctrl_arch}), is our preferred candidate for online adaptation in a real scenario.
However, the analysis is out of scope for this study. More experiments are needed to evaluate stability and performance of the online adaptation process.

\appendix

\subsubsection*{Acknowledgments}
\label{sec:ack}

This work was funded by the European Aerospace Agency under the GSTP programme, activity GT1I-602SA ``Artificial Intelligence techniques for spacecraft attitude control and estimation'' (project acronym: AI4AOCS), contract number 4000145154/24/NL/MGu, lead by Airbus Defence and Space GmbH.
We thank Octavio Arriaga for feedback on our implementation of MR.Q, Shubham Vyas for feedback on the manuscript, and Arthur de Freitas Precht for technical support for the interface to the simulation software.
This work was partially supported by the German Federal Ministry of Research, Technology and Space (BMFTR) under the Robotics Institute Germany~(RIG).


\bibliography{main}
\bibliographystyle{rlj}

\beginSupplementaryMaterials

\section{Action Space}
\label{sec:app_actions}

\begin{longtable}{p{3.5cm}p{6cm}ll}
\toprule
{\bf Action} & {\bf Description} & {\bf Unit} & {\bf Limits}\\
\midrule
\multicolumn{4}{c}{\bf Control Commands (Only in Pure RL or Additive Hybrid Control Mode)}\\
\midrule
$\Delta\delta_{e,\text{cmd}}(t)$ & Change ($\delta_{e,\text{cmd}}(t) - \delta_{e,\text{cmd}}(t-1)$) of symmetric deflection (both flaps trailing edge down). & rad & $\left[-\frac{15}{14}^\circ, \frac{15}{14}^\circ\right]$\\
$\Delta\delta_{a,\text{cmd}}(t)$ & Change of antisymmetric deflection of both flaps. Right flap is deflected trailing edge down, left flap is deflected trailing edge up. & rad & $\left[-\frac{15}{14}^\circ, \frac{15}{14}^\circ\right]$\\
$\tau_{z}(t)$ & The available four thrusters are fired in a group that produces a (yawing) torque about the body-fixed vertical (z) axis. The command is a torque demand that is converted to thruster opening times by low-level functions. & N & $\left[-300, 300\right]$\\
\midrule
\multicolumn{4}{c}{\bf PID Gains (Only in Gain-Scheduling Hybrid Control Mode)}\\
\midrule
$f_{k_{p,\alpha}}$ & Factor for proportional gain of angle of attack error. & decibel & $\left[-6, 6\right]$ \\
$f_{k_{i,\alpha}}$ & Factor for integral gain of angle of attack error. & decibel & $\left[-6, 6\right]$ \\
$f_{k_{d,\alpha}}$ & Factor for derivative gain of angle of attack error. & decibel & $\left[-6, 6\right]$ \\
$f_{k_{p,\beta,\beta}}$ & Factor for proportional gain of yaw angle error to thruster command. & decibel & $\left[-6, 6\right]$ \\
$f_{k_{d,\beta,\beta}}$ & Factor for derivative gain of yaw angle error to thruster command. & decibel & $\left[-6, 6\right]$ \\
$f_{k_{p,\beta,\mu}}$ & Factor for proportional gain of bank angle error to thruster command. & decibel & $\left[-6, 6\right]$ \\
$f_{k_{d,\beta,\mu}}$ & Factor for derivative gain of bank angle error to thruster command. & decibel & $\left[-6, 6\right]$ \\
$f_{k_{p,\mu,\mu}}$ & Factor for proportional gain of bank angle error to flap command. & decibel & $\left[-6, 6\right]$ \\
$f_{k_{i,\mu,\mu}}$ & Factor for integral gain of bank angle error to flap command. & decibel & $\left[-6, 6\right]$ \\
$f_{k_{d,\mu,\mu}}$ & Factor for derivative gain of bank angle error to flap command. & decibel & $\left[-6, 6\right]$ \\
$f_{k_{p,\mu,\beta}}$ & Factor for proportional gain of yaw angle error to flap command. & decibel & $\left[-6, 6\right]$ \\
$f_{k_{d,\mu,\beta}}$ & Factor for derivative gain of yaw angle error to flap command. & decibel & $\left[-6, 6\right]$ \\
\bottomrule
\end{longtable}

\clearpage

\section{Observation Space}
\label{sec:app_observations}

\begin{longtable}{p{3.5cm}p{6cm}ll}
\toprule
{\bf Observation} & {\bf Description} & {\bf Unit} & {\bf Limits}\\
\midrule
\multicolumn{4}{c}{\bf Environment Conditions}\\
\midrule
Altitude & Altitude above the WGS84 geoid. & m & $\left[0, 150k\right]$\\
Mach number $M\!a$ & Velocity as ratio of absolute speed and local speed of sound & unitless & $\left[0, 35\right]$\\
Dynamic pressure $\bar{q}$ & Dynamic pressure depends on air speed $V_a$ and air density $\rho$ through $0.5 V_a^2 \rho$ & Pa & $\left[0, 10k\right]$\\
\midrule
\multicolumn{4}{c}{\bf Commanded Aerodynamic Angles}\\
\midrule
\multicolumn{4}{c}{%
\begin{minipage}{0.9\textwidth}
The commands are generated using a model of the desired response characteristics (second-order transfer function) in the setpoint generator. Tracking the aerodynamic angles is the primary control objective.
\end{minipage}%
}\\
\midrule
$\alpha_{\text{cmd}}(t)$ (angle of attack) & Requested angle between the direction of flight relative to air and the body-fixed frame in the vertical plane. & rad & $\left[-\frac{\pi}{2}, \frac{\pi}{2}\right]$ \\
$\beta_{\text{cmd}}(t)$ (sideslip angle) & Requested angle between the direction of flight relative to air and the body-fixed frame in the horizontal plane. The command can be assumed to be zero all of the time. & rad & $\left[-\frac{\pi}{2}, \frac{\pi}{2}\right]$ \\
$\mu_{\text{cmd}}(t)$ (bank angle) & Requested angle between the direction of lift and the vertical plane around the direction of flight relative to air. & rad & $\left[-\frac{\pi}{2}, \frac{\pi}{2}\right]$ \\
\midrule
\multicolumn{4}{c}{\bf Measured Angles and Angular Rates}\\
\midrule
$\alpha(t)$ (angle of attack) & Measured angle of attack. & rad & $\left[-\frac{\pi}{2}, \frac{\pi}{2}\right]$ \\
$\beta(t)$ (sideslip angle) & Measured angle of sideslip. & rad & $\left[-\frac{\pi}{2}, \frac{\pi}{2}\right]$ \\
$\mu(t)$ (bank angle) & Measured bank angle. & rad & $\left[-\frac{\pi}{2}, \frac{\pi}{2}\right]$ \\
$\alpha(t-1)$ & Last measured angle of attack. & rad & $\left[-\frac{\pi}{2}, \frac{\pi}{2}\right]$ \\
$\beta(t-1)$ & Last measured angle of sideslip. & rad & $\left[-\frac{\pi}{2}, \frac{\pi}{2}\right]$ \\
$\mu(t-1)$ & Last measured bank angle. & rad & $\left[-\frac{\pi}{2}, \frac{\pi}{2}\right]$ \\
$\dot{\phi}(t)$ (roll rate) & Angular velocity about the forward (x) axis of the vehicle. & rad/s & $\left[-10, 10\right]$\\
$\dot{\theta}(t)$ (pitch rate) & Rate of change of the pitch angle. Corresponds to angular velocity about the intermediate y' axis in a standard Euler angle zyx-rotation sequence. & rad/s & $\left[-10, 10\right]$\\
$\dot{\psi}(t)$ (yaw rate) & Rate of change of the yaw angle. Corresponds to angular velocity about the geodetic z axis in a standard Euler angle zyx-rotation sequence. & rad/s & $\left[-10, 10\right]$\\
\midrule
\multicolumn{4}{c}{\bf Last Commands}\\
\midrule
$\delta_{\text{e},\text{cmd}}(t-1)$ & Flap command from last step. See actions for details. & rad & $\left[-\frac{\pi}{2}, \frac{\pi}{2}\right]$ \\
$\delta_{\text{a},\text{cmd}}(t-1)$ & Flap command from last step. See actions for details. & rad & $\left[-\frac{\pi}{2}, \frac{\pi}{2}\right]$ \\
\midrule
\multicolumn{4}{c}{\bf Baseline Control Commands (Only in Hybrid Control Mode)}\\
\midrule
$\delta_{\text{e},\text{basecmd}}(t)$ & Flap command of baseline controller. See actions for details. & rad & $\left[-\frac{\pi}{2}, \frac{\pi}{2}\right]$\\
$\delta_{\text{a},\text{basecmd}}(t)$ & Flap command of baseline controller. See actions for details. & rad & $\left[-\frac{\pi}{2}, \frac{\pi}{2}\right]$\\
$\tau_{\text{z},\text{basecmd}}(t)$ & Thruster command of baseline controller. See actions for details. & N & $\left[-300, 300\right]$\\
\midrule
\multicolumn{4}{c}{\bf PID Error Components}\\
\midrule
$e_\alpha(t)$ & Proportional error for angle of attack. & rad & $\left[-\pi,\pi\right]$ \\
$e_\beta(t)$ & Proportional error for sideslip angle. & rad & $\left[-\pi,\pi\right]$ \\
$e_\mu(t)$ & Proportional error for bank angle. & rad & $\left[-\pi,\pi\right]$ \\
$\int_0^t e_\alpha(\tau) d\tau$ & Integral error for angle of attack. & \si{\radian\second} &\\
$\int_0^t e_\beta(\tau) d\tau$ & Integral error for sideslip angle. & \si{\radian\second} &\\
$\int_0^t e_\mu(\tau) d\tau$ & Integral error for bank angle. & \si{\radian\second} &\\
$\frac{d e_\alpha(t)}{dt}$ & Derivative error for angle of attack. & \si{\radian\per\second} &\\
$\frac{d e_\beta(t)}{dt}$ & Derivative error for sideslip angle. & \si{\radian\per\second} &\\
$\frac{d e_\mu(t)}{dt}$ & Derivative error for bank angle. & \si{\radian\per\second} &\\
\bottomrule
\end{longtable}

\section{Design of Reward Function}
\label{sec:app_reward}

\subsection{Sensitivity Analysis of Reward Function}

\begin{figure}[tb]
\centering
\begin{subfigure}[c]{\textwidth}
\centering
\includegraphics{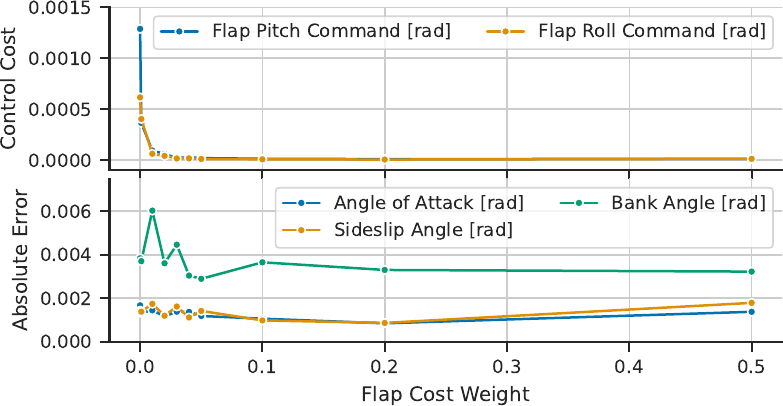}
\subcaption{Influence of flap cost weight on control cost and absolute error of aerodynamic angles.\label{fig:reward_sensitivity_flap_cost}}
\end{subfigure}
\begin{subfigure}[c]{\textwidth}
\centering
\includegraphics{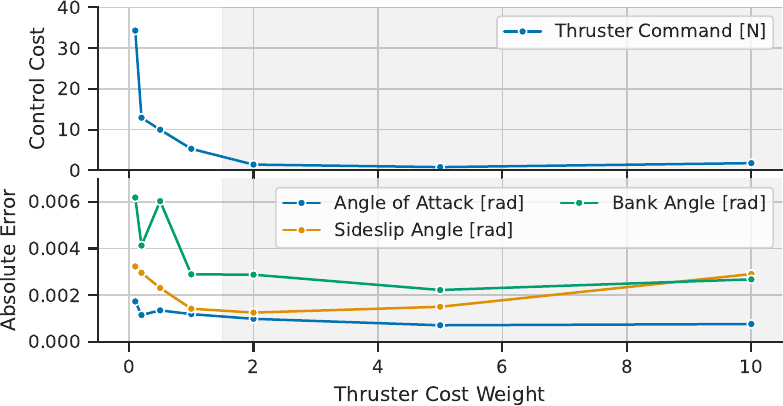}
\subcaption{Influence of thruster cost weight on control cost and absolute error of aerodynamic angles. In the gray region, the policy was not able to follow the complete trajectory.\label{fig:reward_sensitivity_thruster_cost}}
\end{subfigure}

\caption{Sensitivity analysis of for weights in the reward function.}
\label{fig:reward_cost_weights}
\end{figure}

For various configurations of the control cost weights (flap cost weights $w_{\delta_e} = w_{\delta_a}$ and thruster cost weight $w_{\tau_z}$), we report the interquartile mean of absolute errors and control costs of the obtained controllers over ten randomly sampled evaluation trajectories after 2.4M training steps with one training run of MR.Q in Figure \ref{fig:reward_cost_weights}.

We see that increasing the flap cost weights $w_{\delta_e} = w_{\delta_a}$ has the desired effect of reducing the flap control cost without sacrificing tracking performance around a sweet spot. Good policies with low control costs are obtained in the range of $\left[0.03, 0.2\right]$. However, a weight of $0.2$ considerably reduces sample efficiency in comparison to lower values, indicating that it makes the learning task harder. Usually about $250k$ to $600k$ training steps are required to follow the trajectory completely during training. With $w_{\delta_e} = w_{\delta_a} = 0.2$, about $1M$ training steps were required. For a weight of $0.5$, $2.2M$ training steps are required. Although this analysis was done after the experiments under nominal conditions, $w_{\delta_e} = w_{\delta_a} = 0.05$ seems to be a good choice in hindsight as it balances sample efficiency, control cost, and control performance.

Similarly, increasing the thruster cost weight has the desired effect of reducing control cost. However, the errors of the aerodynamic angles decrease at the same time. The best possible weight is about $w_{\tau_z} = 1$ as increasing the weight further makes the learning problem so hard that it is not possible to follow the whole trajectory after $2.5M$ training steps.

\subsection{Monitoring Training}

Since the maximum value of $r_t$ is 1, the maximum value of the expected return approximated by the value function is $\frac{1}{1-\gamma}$ under the discounted infinite horizon model, e.g., for $\gamma=0.99$ the maximum return is 100 if we assume no control costs and that the attitude is tracked perfectly. Hence, we are able to directly track the performance of the policy in comparison to the optimum during training in algorithms based on DDPG (i.e., TD3, TD7, MR.Q), since the loss of the policy is the negative expected return approximated by the value function. The best possible value would be -100 under the condition of perfect approximation.

\subsection{Considered Reward Functions}

In non-systematic, preliminary experiments we explored a wide range of reward functions and components of reward functions. We tried the following:
\begin{itemize}
\item Negative attitude cost: sum of squared differences between commanded and measured aerodynamic angles.
\item Adding a penalty for thruster forces.
\item Adding a penalty for flap angle changes.
\item Adding a penalty for derivative of the aerodynamic angles. We do not use this component anymore because it is often required to change the aerodynamic angles considerably to track the commanded trajectory.
\item Adding an extra penalty for large deviations from commanded angles (unsafe state penalty).
\item Terminating an episode when the orientation is out of the safe range and adding a large penalty of -1000 or -8000 (stop penalty).
\item Adding a reward for each step that the episode is not aborted (healthy reward), which results in significantly longer simulations, however, sometimes at the cost of less accurate tracking of the orientation.
\item Replacing angle cost by forward reward, which rewards moving closer to the commanded attitude and is inspired by the reward function of the MuJoCo environments from Gymnasium.
\item Initially, we set the weight of the control cost components to 0.001 and then increased it to 0.01, 0.1, 0.5, and 1. Higher weights lead to less switching between flap angles, less extensive use of thrusters, and generally less oscillation of commands. The current individual weights are a good compromise between tracking performance and avoiding excessive overcompensation of errors.
\item We tried to remove the healthy reward and use a terminal cost instead, that rewards low altitude of the vehicle when the episode terminates. This did not lead to better results, even when evaluating policies obtained with the previous reward under the new reward.
\item We tried to remove the healthy reward and use an attitude that is always positive and more positive when the attitude cost is lower. We call this the attitude reward and it is included in the final version of the reward function. The idea behind this is to substitute the healthy reward and the attitude cost component, since it is always better to survive longer and it is better to have a low attitude cost. The approach was successful, so we do not use any attitude cost, terminal cost, or stop penalty anymore.
\end{itemize}

\section{Simulation Details}
\label{app:sim_details}

\subsection{Common Configuration}

\paragraph{Initial conditions:}
The vehicle starts at the outer atmosphere of Earth with a fixed altitude of about \SI{93}{\kilo\meter} and fixed velocity of approximately Mach 26.75. The dynamic pressure is low with approximately \SI{54}{\pascal} in this state.

\paragraph{Flight path:}
To avoid overfitting to a specific flight path, we randomize the trajectory given by the guidance system in each episode by setting two parameters $\gamma$ and $\Delta \chi_{\max}$ (see Table \ref{tab:env_config}). The flight path angle $\gamma$ determines the desired speed of descent. The maximum deviation from the flight course $\Delta \chi_{\max}$ controls bank angle reversals. These reversals are necessary because the bank angle of the vehicle influences not just the speed of descent, but also the direction of the flight. In order to follow a straight path, the bank angle has be reversed when $\Delta \chi$ exceeds the given threshold.
This has the effect of modifying the environment dynamics and reward function from the perspective of the agent because it modifies the commanded aerodynamic angles that are part of the state space as well as the PID error components that are part of the observation space.

\begin{table}[t]
\caption{Configuration of simulation.\label{tab:env_config}}
\centering
\begin{tabular}{llll}
\toprule
\thead{Variable} & \thead{Nominal} & \thead{Distribution (Nominal)} & \thead{Distribution (Dynamics Randomization)} \\
\midrule
\multicolumn{4}{c}{\thead{Trajectory Parameters}}\\
\midrule
$\gamma$ & \ang{-1.0} & $\mathcal{U}(-1.1^\circ, -0.9^\circ)$ & $\mathcal{U}(-1.1^\circ, -0.9^\circ)$ \\
$\Delta \chi_{\max}$ & \ang{3.25} & $\mathcal{U}(1.5^\circ, 5^\circ)$ & $\mathcal{U}(1.5^\circ, 5^\circ)$ \\
\midrule
\multicolumn{4}{c}{\thead{Initial Attitude}}\\
\midrule
$\alpha(0)$ & \ang{45.024} & constant & $\mathcal{U}(\alpha_{0,\text{nominal}} - 5^\circ, \alpha_{0,\text{nominal}} + 5^\circ)$ \\
$\beta(0)$ & \ang{0.046} & constant & $\mathcal{U}(\beta_{0,\text{nominal}} - 5^\circ, \beta_{0,\text{nominal}} + 5^\circ)$ \\
$\mu(0)$ & \ang{61.141} & constant & $\mathcal{U}(\mu_{0,\text{nominal}} - 5^\circ, \mu_{0,\text{nominal}} + 5^\circ)$ \\
\midrule
\multicolumn{4}{c}{\thead{Inertia}}\\
\midrule
$m_0$ & \SI{1640}{\kilo\gram} & constant & context, $m_0 \in \left[ 1312, 1968 \right]$ \si{\kilo\gram} \\
$\boldsymbol{I}$ & $\text{diag}\left(\begin{array}{c}
492 \\
2247 \\
2358 \\
\end{array}\right) \text{kg} \cdot \text{m}^2$
& constant & context, see main text \\
\midrule
\multicolumn{4}{c}{\thead{Actuation}}\\
\midrule
$\omega_0$ & \SI{30}{\radian\per\second} & constant & context, $\omega_0 \in \left[12, 30\right]$ \si{\radian\per\second} \\
\bottomrule
\end{tabular}
\end{table}

\subsection{Nominal Conditions}

Under nominal conditions, the initial attitude, vehicle mass, the vehicle's inertia tensor, and the actuation are constant as summarized in Table \ref{tab:env_config}.

\subsection{Dynamics Randomization}

In comparison to nominal conditions, we additionally randomize the initial attitude for each episode according to Table \ref{tab:env_config} and the context vector contains the vehicle's mass, inertia tensor, and flap actuator bandwidth.

To model uncertainty in the actuation of the flaps, we modify the flap actuator bandwidth $\omega_0$. Lower values make controlling the vehicle harder. The baseline controller becomes unstable for values under $14$ \si{\radian\per\second}.

The initial mass is varied in a predefined range according to Table \ref{tab:env_config}. The vehicle's mass defines its gravitational force and, hence, influences the evolution of the flight path. The flight path then defines the commanded aerodynamic angles, as it is required to increase or decrease the lift to compensate for the change in mass.

To generate physically plausible inertia tensors, we define perturbations of the nominal inertia tensor with the following procedure. Since the nominal inertia tensor is diagonal, we apply a perturbation directly to the principal moments and then rotate the inertia tensor.
To modify the principal moment, we use a vector $\boldsymbol{a} \in \left[-0.1, 0.1\right]^3,$
and to modify the rotation, we use a rotation vector $\boldsymbol{\omega} = \theta \hat{\boldsymbol{\omega}} \in \mathbb{R}^3$ with an angle
$\theta \in \left[-10^\circ, 10^\circ\right]$ and axis
$\hat{\boldsymbol{\omega}} \in S^2 \subset \mathbb{R}^3$.
With the exponential map $\text{Exp}: \mathbb{R}^3 \rightarrow SO(3)$ \citep{Sola2018}, we compute the rotation matrix from the rotation vector so that the nominal inertia is perturbed with
\[
\boldsymbol{I} = \text{Exp}(\boldsymbol{\omega})
\cdot
\text{diag}(\boldsymbol{1}_3 + \boldsymbol{a})
\cdot
\boldsymbol{I}_{\text{nominal}}
\cdot
\text{Exp}(\boldsymbol{\omega})^T,
\]
where $\text{diag}(\boldsymbol{x})$ transforms a vector $\boldsymbol{x} \in \mathbb{R}^3$ to a diagonal matrix.
We generate a random set of rotation axes by sampling from a 3D standard normal distribution $\mathcal{N}(\boldsymbol{0}_3, \boldsymbol{I}_{3\times 3})$ and normalizing the vector.

\clearpage

\section{Hyperparameters of RL Algorithms}
\label{sec:app_hyperparams}

\begin{longtable}{cp{4.5cm}p{1.5cm}p{1.5cm}p{1.5cm}p{1.5cm}}
\toprule
& Hyperparameter & MR.Q & TD7 & TD3 & SAC \\
\midrule
\multirow{5}{*}{\rotatebox{90}{Common}}
& Discount factor~$\gamma$      & $0.99$ & $0.99$ & $0.99$ & $0.99$ \\
& Replay buffer capacity    & $2$M & $2$M & $2$M & $2$M \\
& Mini-batch size           & $256$ & $256$ & $256$ & $256$ \\
& Target update frequency~$T_\text{target}$   & $250$ & $250$ & - & -\\
& Target update rate~$\tau$ & - & - &  $5\cdot 10^{-3}$ &$5\cdot 10^{-3}$ \\
\midrule
\multirow{6}{*}{\rotatebox{90}{MR.Q}} & Dynamics loss weight $\lambda_\text{Dynamics}$ & $1$ & - & - & - \\
& Reward loss weight $\lambda_\text{Reward}$ & $0.1$ & - & - & - \\
& Terminal loss weight $\lambda_\text{Terminal}$ & $0.1$ & - & - & - \\
& Activation loss weight $\lambda_\text{pre-activ}$ & $1\text{e}-5$ & - & - & - \\
& Encoder horizon $H_\text{Enc}$ & $5$ & - & - & - \\
& Multi-step returns horizon $H_Q$ & $3$ & - & - & - \\
\midrule
\multirow{5}{*}{\rotatebox{90}{TD3}}
& Target policy noise~$\sigma$      & $\mathcal{N}(0,0.2^2)$ & $\mathcal{N}(0,0.2^2)$ & $\mathcal{N}(0,0.2^2)$ & - \\
& Target policy noise clipping~$c$  & $(-0.3, 0.3)$ & $(-0.5, 0.5)$ & $(-0.5, 0.5)$ & - \\
& Initial random exploration steps & $10$k & $25$k & $25$k & $5$k \\
& Exploration noise    & $\mathcal{N}(0,0.2^2)$ & $\mathcal{N}(0,0.1^2)$ & $\mathcal{N}(0,0.2^2)$ & - \\
& Policy delay              & $1$ & $2$ & $2$ & $1$ \\
\midrule
\multirow{2}{*}{\rotatebox{90}{LAP}}
& Probability smoothing $\alpha$    & $0.4$ & $0.4$ & - & - \\
& Minimum priority                  & $1$ & $1$ & - & - \\
\midrule
\multirow{7}{*}{\rotatebox{90}{Value Network}}
& Optimizer        & AdamW\footnote{\citet{loshchilov2018decoupled}} & Adam\footnote{\citet{Kingma2014_adam}} & Adam & Adam \\
& Learning rate    & $3\text{e}-4$ & $3\text{e}-4$ & $3\text{e}-4$ & $1\text{e}-3$ \\
& Hidden dim & $512$ & $256$ & $256$ & $256$ \\
& Activation function & ELU\footnote{\citet{clevert2015fast}} & ELU & ReLU & ReLU \\
& Weight initialization & Xavier uniform\footnote{\citet{Glorot2010understanding}} & LeCun normal\footnote{\citet{Klambauer2017}} & LeCun normal & LeCun normal \\
& Bias initialization & $0$ & $0$ & $0$ & $0$ \\
& Gradient clip norm & $20$ & - & - & - \\
\midrule
\multirow{7}{*}{\rotatebox{90}{Policy Network}}
& Optimizer        & AdamW & Adam & Adam & Adam \\
& Learning rate    & $3\text{e}-4$ & $3\text{e}-4$ & $3\text{e}-4$ & $3\text{e}-4$ \\
& Hidden dim & $512$ & $256$ & $256$ & $256$ \\
& Activation function & ReLU\footnote{\citet{Glorot2011_relu}} & ReLU & ReLU & Swish\footnote{\citet{Ramachandran2017}} \\
& Weight initialization & Xavier uniform & LeCun normal & LeCun normal & LeCun normal \\
& Bias initialization & $0$ & $0$ & $0$ & $0$ \\
\midrule
\multirow{8}{*}{\rotatebox{90}{Encoder Network}}
& Optimizer        & AdamW & Adam & - & - \\
& Learning rate    & $1\text{e}-4$ & $3\text{e}-4$ & - & - \\
& Weight decay     & $1\text{e}-4$ & - & - & - \\
& $\mathbf{z}_s$ dim & $512$ & $256$ & - & - \\
& $\mathbf{z}_{sa}$ dim & $512$ & $256$ & - & - \\
& $\mathbf{z}_a$ dim & $256$ & $256$ & - & - \\
& Hidden dim & $512$ & $256$ & - & - \\
& Activation function & ELU & ELU & - & - \\
\multirow{7}{*}{\rotatebox{90}{Encoder (cont.)}} & Weight initialization & Xavier uniform & LeCun normal & - & -
 \\
& Bias initialization & $0$ & $0$ & - & - \\
& Reward bins & $65$ & - & - & - \\
& Reward range & $[-10,10]$ (effective: $[-22\text{k}, 22\text{k}]$) & - & - & - \\
\midrule
\multirow{2}{*}{\rotatebox{90}{SAC}}  & Entropy optimizer & - & - & - & Adam \\
& Entropy learning rate     & - & - & - & $1\text{e}-3$ \\
\bottomrule
\end{longtable}

While the MR.Q algorithm is a theoretically sound approach \citep{fujimoto2025_mrq}, we had to modify it slightly in comparison to its original implementation.
In the original implementation, the dynamics loss compares the latent state computed from unrolled dynamics $z_{sa}^T W_p \in \mathbb{R}$ to the latent target state obtained by applying the state encoder $f_\omega$ to a state $s'$ from the replay buffer $f_\omega(s') = z_{s'}$, in which the state encoder applies layer normalization and ELU, hence, limits the output to $z_{s'} \in \mathbb{R}_{\geq -1}$.
Initially, we had this as a bug in our implementation of MR.Q, but we found that removing ELU from the last layer of the state encoder enhances performance.
Hence, our implementation does apply layer normalization without ELU after the last layer of the state encoder, which resembles the behavior of the state encoder in TD7 \citep{Fujimoto2023_td7}, in which AvgL1Norm is supposed to protect from monotonic growth of the features.

\section{Hyperparameters of Task Scheduling Algorithms}
\label{sec:app_ts_hyperparams}

For task scheduling, we sample a discrete set of contexts and create one replay buffer per context. For each update, we first randomly sample the replay buffer and then a batch of samples from this replay buffer. This ensures that we continue training each task even though a task is not selected anymore.

In comparison to the original SMT \citep{Cho2024}, we do not perform any network resets, which does not seem to be necessary in MR.Q. Furthermore, we do not learn a task encoding.

\begin{longtable}{clll}
\toprule
& Hyperparameter & Active MT & SMT \\
\midrule
\multirow{5}{*}{\rotatebox{90}{Common}}
& RL algorithm & MR.Q & MR.Q \\
& Number of tasks $|\mathcal{T}|$ & 50 & 50 \\
& Training steps & $15M$ & $15M$ \\
& Scheduling interval & 1 episode & 1 episode \\
& Replay buffer size per task & $\max\left(200k, \frac{2M}{|\mathcal{T}|}\right)$ & $\max\left(200k, \frac{2M}{|\mathcal{T}|}\right)$\\
\midrule
\multirow{3}{*}{\rotatebox{90}{D-UCB}}
& Upper bound for task selection reward $r_{\max}$ & $10.8k$ & - \\
& Discount factor $\gamma_{\mathcal{T}}$ & $0.95$ & - \\
& Padding function strength $\xi$ & $1\text{e}-4$ & - \\
\midrule
\multirow{7}{*}{\rotatebox{90}{SMT}}
& Stage 1 budget $B_1$ & - & $12.75M$ \\
& Stage 2 budget $B_2$ & - & $2.25M$ \\
& $\kappa$ & - & $0.7$ \\
& $K$ & - & $8$ \\
& Threshold for unsolved tasks $m$ & - & $-100$\\
& Threshold for solved tasks $M$ & - & $10k$\\
& Reset interval & - & No resets \\
\bottomrule
\end{longtable}

\clearpage

\section{Ablation Studies}
\label{app:ablations}

We perform ablation studies to find out why MR.Q works in this application.
To evaluate the effect of algorithm components and hyperparameters in the MR.Q algorithm, we train for $2.4M$ steps in the single task setting with one seed, and measure the accumulated reward over ten evaluation episodes. Although the results of these studies are not statistically robust with only one training seed, we expect the conclusions to be correct, since MR.Q is a robust algorithm with mostly consistent behavior across random seeds and the results are unambiguous.

\subsection{Effect of Policy Activation Penalty}

\scalebox{0.98}{
\begin{tabular}{llllll}
\toprule
& \multicolumn{4}{c}{Activation loss weight $\lambda_\text{pre-activ}$}\\
\cmidrule(lr){2-5}
& 0 & $10^{-6}$ & $10^{-5}$ (default) & $10^{-4}$ \\
\midrule
Metric & \multicolumn{4}{c}{IQM $\pm$ interquartile std. dev.}\\
\midrule
Return                & 10587 $\pm$ 166     & 10467 $\pm$ 163     & 10492 $\pm$ 181     & 10570 $\pm$ 193\\
Flap Roll Cmd. [rad]  & 1.6e-5 $\pm$ 3.1e-5 & 1.6e-5 $\pm$ 10e-5  & 2.0e-5 $\pm$ 5.0e-5 & 2.1e-5 $\pm$ 6.0e-5 \\
Flap Pitch Cmd. [rad] & 0.8e-6 $\pm$ 2.5e-5 & 1.4e-5 $\pm$ 8.7e-5 & 1.2e-5 $\pm$ 3.7e-5 & 1.0e-5 $\pm$ 4.7e-5 \\
Thruster Cmd. [N]     & 2.97 $\pm$ 3.03     & 4.31 $\pm$ 3.6      & 5.29 $\pm$ 4.69     & 4.60 $\pm$ 4.42 \\
\bottomrule
\end{tabular}
}

Similarly to the control cost of the reward function, we expected that the activation loss of the policy forces the policy to reduce the magnitude of control commands. Other algorithms do not have this component and learn policies that often result in bang-bang control. However, setting $\lambda_\text{pre-activ}=0$ does not decrease the performance. The activation penalty is not contributing to the considerable performance improvement of MR.Q in comparison to TD3 and TD7.

\subsection{Effect of Model-Based Representation Learning}

\begin{longtable}{lllll}
\toprule
Encoder loss & $\lambda_\text{Dynamics}$ & $\lambda_\text{Reward}$ & $\lambda_\text{Terminal}$ & Return (IQM $\pm$ interquartile std. dev.) \\
\midrule
No loss & 0 & 0 & 0 & 2987$\pm$277 \\
No reward loss & 1 & 0 & 0.1 & 10455$\pm$212 \\
No dynamics loss & 0 & 0.1 & 0.1 & 10630$\pm$193 \\
Default & 1 & 0.1 & 0.1 & 10492$\pm$181 \\
\bottomrule
\end{longtable}

The model-based representation learned by the encoder is the main difference of MR.Q in comparison to TD3 and TD7. Deactivating the loss for the encoder completely, i.e., using random features, results in a performance similar to TD3 and TD7, which confirms that the model-based representation of MR.Q is important for the application. Interestingly, deactivating either the dynamics loss or the reward loss has a negligible effect on the final performance. Note that these results are specific for this application, in which a useful and meaningful reward can be computed in every step. The behavior might be different for environments with sparse reward.

\subsection{Effect of Horizon for Multi-Step Returns in Critic Loss}

\begin{longtable}{lllll}
\toprule
Multi-step returns horizon $H_Q$ & 1 & 3 (default) & 5 & 10 \\
\midrule
Return (IQM) & 3184 & 10492 & 9250 & 4483 \\
\bottomrule
\end{longtable}

As EVE-TD3 \citep{liu_attitude_2022} uses $H_Q=30$, we are interested in the effect of the hyperparameter in this application with MR.Q. The horizon controls the bias-variance trade-off in the target value for the action-value function (see also \citet{Schulman2016} for a discussion in the context of advantage function estimation).
The results demonstrate that the multi-step return target for the critic is a main reason for the good performance of MR.Q in this application.
Furthermore, MR.Q's default value of $H_Q=3$ is also an optimum in this application.
Considering the control frequencies of \SI{14}{\hertz} in our controller and \SI{50}{\hertz} in the controller of \citet{liu_attitude_2022} (control step size of \SI{20}{\milli\second}), we use a horizon of ca. \SI{0.214}{\second} and \citet{liu_attitude_2022} use a horizon of \SI{0.6}{\second} in simulation time, which is in the same order of magnitude.

\section{Code}

Although we cannot make the code for the simulation and experiments publicly available, the source code for all reinforcement learning algorithms used in our experiments is available at \url{https://github.com/mlaux1/rl-blox} (version 0.5.7, JAX version 0.7.0, flax version 0.11.1).

For statistical comparison of RL algorithms, we rely on the open-source library rliable of \citet{Agarwal2021_statistical_precipe} available at \url{https://github.com/google-research/rliable}.

\section{Computation Time}

One run with 10 million time steps in the single task setting takes about 48 h with MR.Q and TD7, 24 h with TD3, and 30 h with SAC. In total, we analyzed 10 training seeds and two control architectures (pure RL and additive hybrid) for each algorithm, resulting in $(48 + 48 + 24 + 30) \times 10 \times 2 = 3000$ hours of wall-clock training time to generate the learning curves. However, the runs could be parallelized. In addition, we used another script to run the performance evaluation of the stored checkpoints.

\clearpage

\section{Application-Specific Performance Analysis Under Nominal Conditions}
\label{app:performance_single_task}

\subsection{Tracking Performance Over Full Trajectory}

\begin{figure}[h!]
\centering
\includegraphics[width=0.85\linewidth]{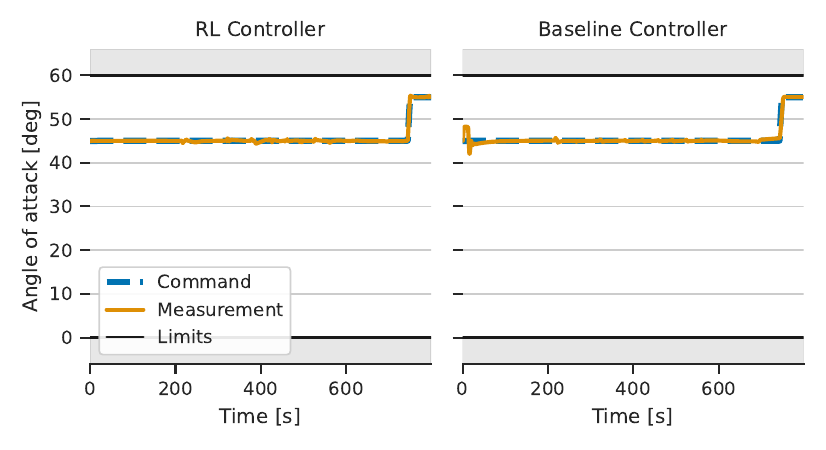}
\includegraphics[width=0.85\linewidth]{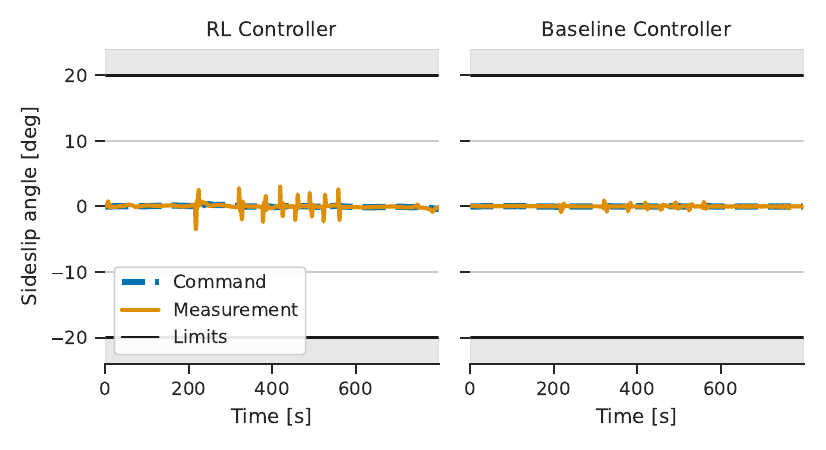}
\includegraphics[width=0.85\linewidth]{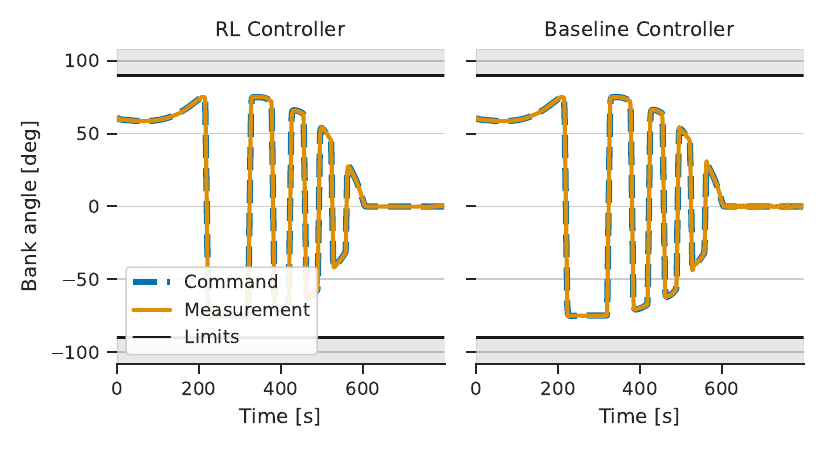}
\caption{Comparison of commanded and measured aerodynamic angles for pure RL controller based on MR.Q trained under nominal conditions and baseline controller.}
\label{fig:st_tracking}
\end{figure}

\clearpage

\subsection{Tracking Performance at Critical Points}

\begin{figure}[h!]
\centering
\includegraphics[width=0.85\linewidth]{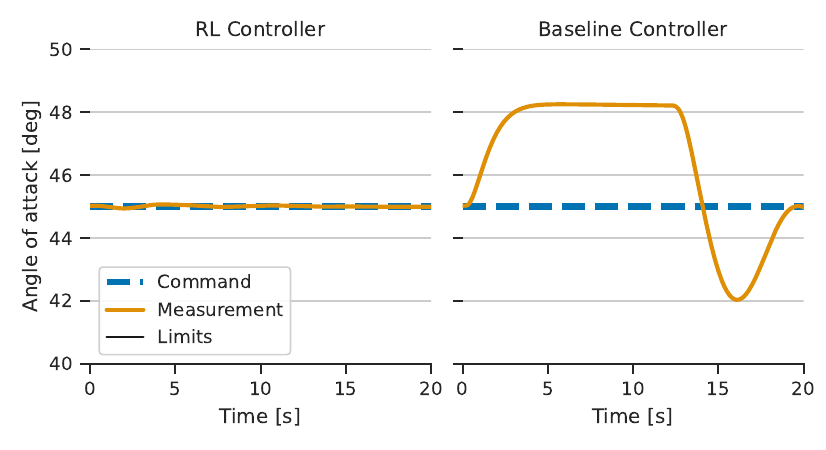}
\includegraphics[width=0.85\linewidth]{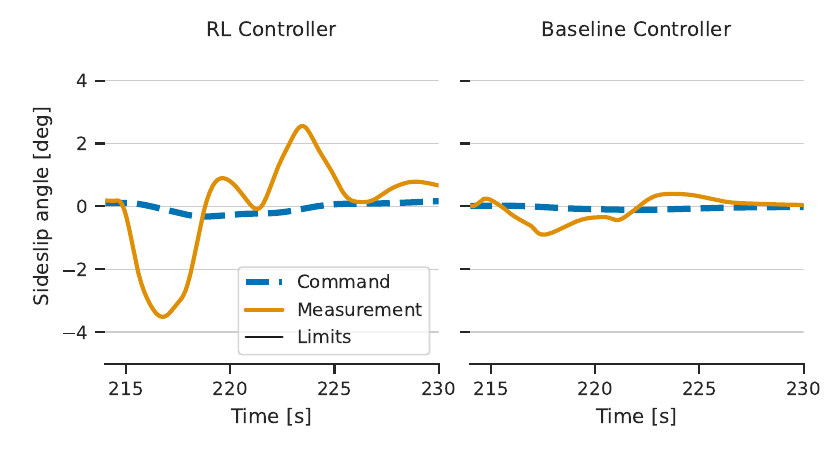}
\includegraphics[width=0.85\linewidth]{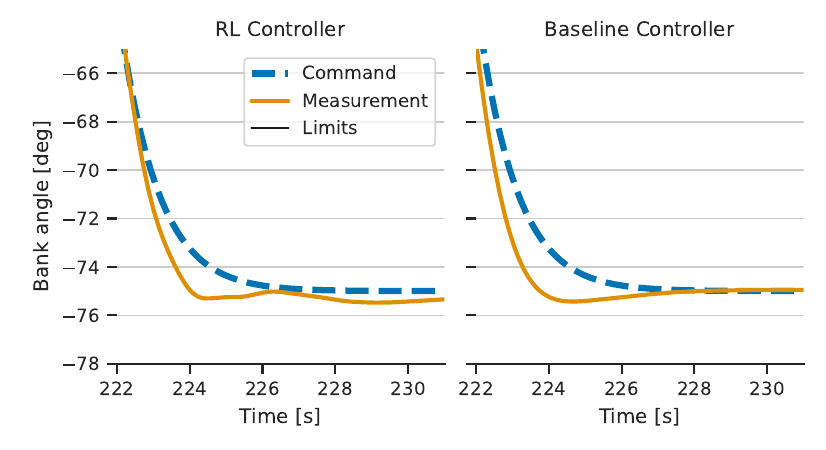}
\caption{Comparison of commanded and measured aerodynamic angles for pure RL controller based on MR.Q trained under nominal conditions and baseline controller. Zoomed in at critical points of the trajectory.}
\label{fig:st_tracking_zoom}
\end{figure}

\clearpage

\subsection{Commands}

\begin{figure}[h!]
\centering
\includegraphics[width=0.85\linewidth]{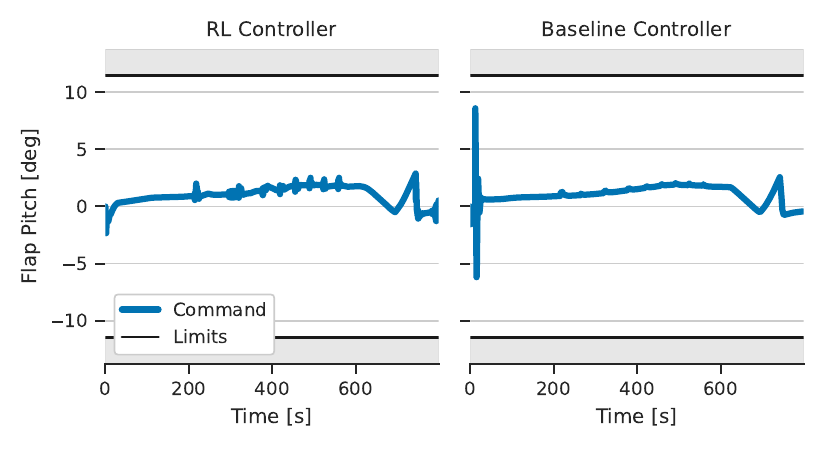}
\includegraphics[width=0.85\linewidth]{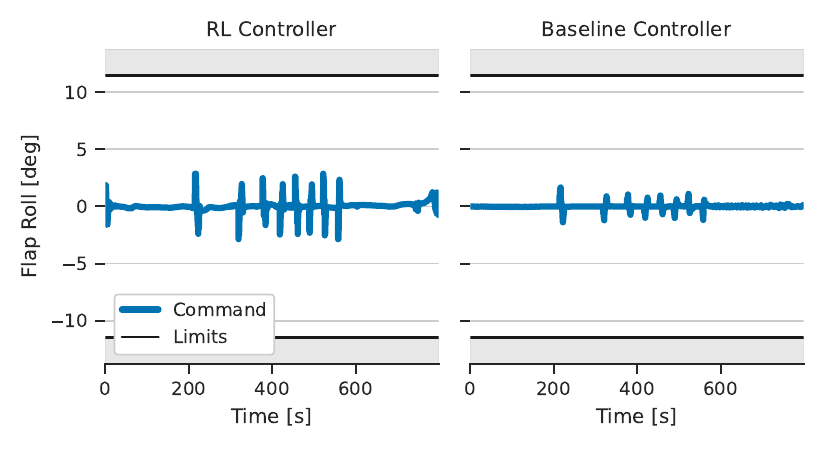}
\includegraphics[width=0.85\linewidth]{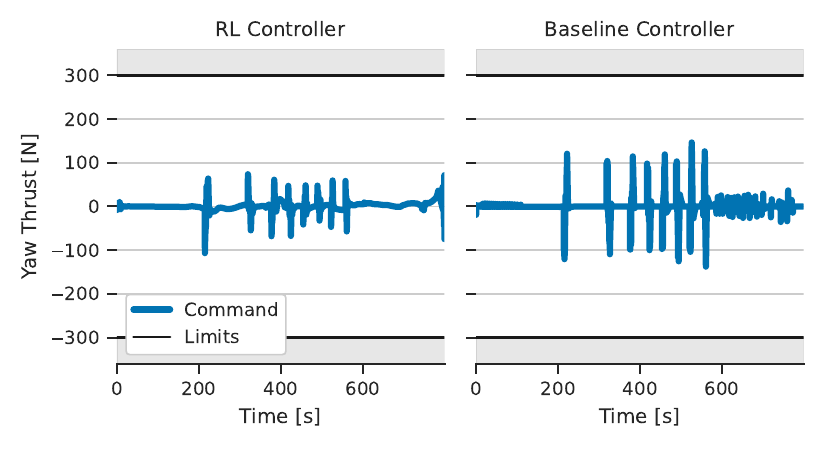}
\caption{Control commands for pure RL controller based on MR.Q trained under nominal conditions and baseline controller.}
\label{fig:st_control}
\end{figure}

\clearpage

\subsection{Error Distribution}

\begin{figure}[h!]
\centering
\includegraphics[width=0.55\linewidth]{st/dist_angle_of_attack_error}
\includegraphics[width=0.55\linewidth]{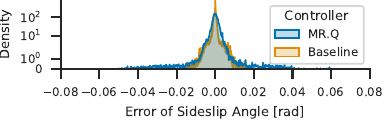}
\includegraphics[width=0.55\linewidth]{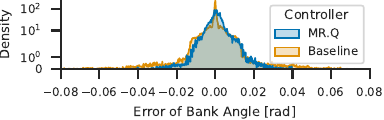}
\caption{Distributions of aerodynamic angle errors. The results are obtained for the best policy and 10 evaluation seeds. A kernel density estimation was applied to the sampled errors. Note that density is plotted on a logarithmic scale to highlight the tails. The peak of the baseline controller around $0$ is more pronounced on a linear scale.}
\label{fig:st_dist_error_zoom}
\end{figure}

\subsection{Distribution of Control Cost}

\begin{figure}[h!]
\centering
\includegraphics[width=0.55\linewidth]{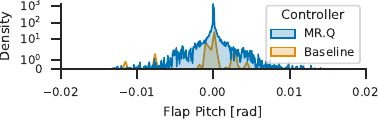}
\includegraphics[width=0.55\linewidth]{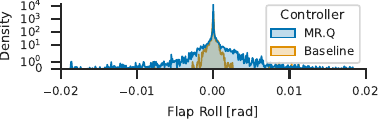}
\includegraphics[width=0.55\linewidth]{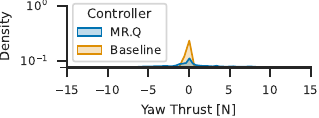}
\caption{Distributions of control commands. The results are obtained for the best policy and 10 evaluation seeds. A kernel density estimation was applied to the sampled control commands. Note that density is plotted on a logarithmic scale to highlight the tails. The peak of the baseline controller around $0$ is more pronounced on a linear scale.
}
\label{fig:st_dist_control}
\end{figure}

\clearpage

\section{Error Percentiles After Training With Dynamics Randomization}
\label{app:mt_percentiles}

\begin{figure}[h!]
\centering
\includegraphics{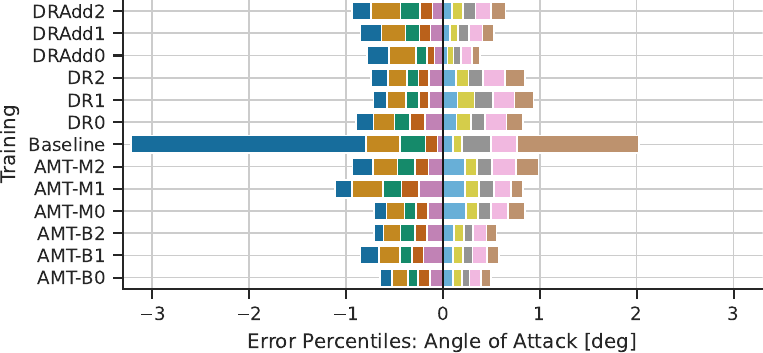}\\[2ex]
\includegraphics{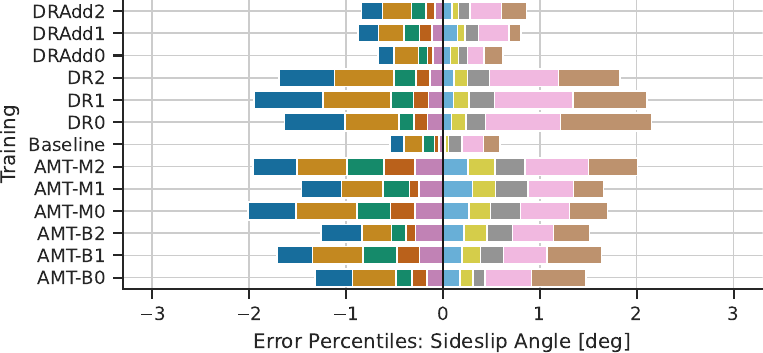}\\[2ex]
\includegraphics{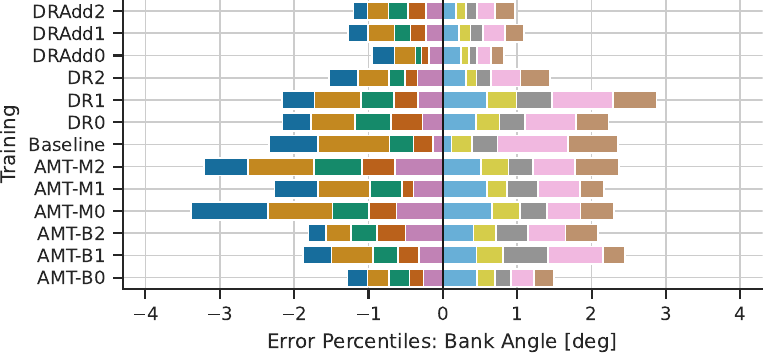}
\caption{Distribution of aerodynamic angle errors evaluated over 100 test contexts. The overlapping bars correspond to the following percentiles of the distribution: 1, 2, 5, 10, 20, 80, 90, 95, 98, 99.}
\end{figure}

\clearpage

\section{Domain-Specific Performance Analysis}
\label{app:mt_metrics}

\begin{table}[h!]
\centering
\caption{Evaluation of robustness of controllers with domain-specific metrics. Cell background color indicates \colorbox{green!40}{best values} or \colorbox{green!15}{values better than the baseline} (only if success rate is $>$0\%). Appendix~\ref{app:mt_percentiles} contains more details on the error distributions.\label{tab:robustness_full}}
{\scriptsize
\begin{longtable}{llr|llll|llll|llll}
\toprule
Training & \rotatebox{90}{Seed} & \rotatebox{90}{Success} & \rotatebox{90}{Median} & \rotatebox{90}{90-Perc.} & \rotatebox{90}{95-Perc.} & \rotatebox{90}{98-Perc.} & \rotatebox{90}{Median} & \rotatebox{90}{90-Perc.} & \rotatebox{90}{95-Perc.} & \rotatebox{90}{98-Perc.} & \rotatebox{90}{Median} & \rotatebox{90}{90-Perc.} & \rotatebox{90}{95-Perc.} & \rotatebox{90}{98-Perc.} \\
\midrule
& & &\multicolumn{4}{c}{$|e_\alpha|$ [deg]}&\multicolumn{4}{c}{$|e_\beta|$ [deg]} &\multicolumn{4}{c}{$|e_\mu|$ [deg]}\\
\midrule
Baseline & - & 79\% & \cellcolor{green!40}0.05 & 0.46 & 0.67 & 2.96 & \cellcolor{green!40}0.02 & \cellcolor{green!40}0.20 & \cellcolor{green!40}0.36 & \cellcolor{green!40}0.57 & \cellcolor{green!40}0.06 & 0.72 & 1.43 & 2.35 \\
\midrule
DR (Add.) & 0 & \cellcolor{green!15}93\% & \cellcolor{green!40}0.05 & \cellcolor{green!40}0.23 & \cellcolor{green!40}0.34 & \cellcolor{green!15}0.59 & 0.07 & 0.25 & 0.41 & 0.65 & 0.17 & \cellcolor{green!40}0.42 & \cellcolor{green!40}0.59 & \cellcolor{green!40}0.87 \\
DR (Add.) & 1 & \cellcolor{green!15}93\% & 0.07 & \cellcolor{green!15}0.37 & \cellcolor{green!15}0.56 & \cellcolor{green!15}0.83 & 0.07 & 0.30 & 0.53 & 0.85 & 0.15 & \cellcolor{green!15}0.60 & \cellcolor{green!15}0.84 & \cellcolor{green!15}1.11 \\
DR (Add.) & 2 & \cellcolor{green!15}94\% & 0.07 & \cellcolor{green!15}0.33 & \cellcolor{green!15}0.47 & \cellcolor{green!15}0.71 & 0.11 & 0.39 & 0.60 & 0.84 & 0.17 & \cellcolor{green!15}0.60 & \cellcolor{green!15}0.84 & \cellcolor{green!15}1.19 \\
\midrule
DR & 0 & \cellcolor{green!40}100\% & 0.12 & 0.47 & \cellcolor{green!15}0.63 & \cellcolor{green!15}0.86 & 0.09 & 0.45 & 0.88 & 1.87 & 0.23 & 1.14 & 1.63 & \cellcolor{green!15}2.19 \\
DR & 1 & \cellcolor{green!15}99\% & 0.10 & \cellcolor{green!15}0.45 & \cellcolor{green!15}0.62 & \cellcolor{green!15}0.84 & 0.10 & 0.53 & 1.06 & 2.03 & 0.34 & 1.31 & 1.82 & 2.57 \\
DR & 2 & \cellcolor{green!15}98\% & 0.10 & \cellcolor{green!15}0.39 & \cellcolor{green!15}0.55 & \cellcolor{green!15}0.80 & 0.09 & 0.49 & 0.95 & 1.76 & 0.25 & \cellcolor{green!15}0.69 & \cellcolor{green!15}0.98 & \cellcolor{green!15}1.49 \\
\midrule
RR & 0 & \cellcolor{green!15}98\% & 0.15 & \cellcolor{green!15}0.45 & \cellcolor{green!15}0.58 & \cellcolor{green!15}0.75 & 0.20 & 0.68 & 0.96 & 1.42 & 0.30 & 1.21 & 1.52 & \cellcolor{green!15}1.98 \\
RR & 1 & \cellcolor{green!15}96\% & 0.25 & 0.71 & 0.82 & \cellcolor{green!15}0.95 & 0.26 & 0.97 & 1.21 & 1.70 & 0.57 & 1.37 & 1.78 & 2.36 \\
RR & 2 & \cellcolor{green!15}95\% & 0.10 & \cellcolor{green!15}0.33 & \cellcolor{green!15}0.43 & \cellcolor{green!40}0.58 & 0.15 & 0.46 & 0.73 & 1.27 & 0.26 & 0.87 & \cellcolor{green!15}1.11 & \cellcolor{green!15}1.35 \\
\midrule
SMT & 0 & 0\% & 0.05 & 0.27 & 0.43 & 0.67 & 0.23 & 0.76 & 1.24 & 1.80 & 0.40 & 0.94 & 1.48 & 2.57 \\
SMT & 1 & 0\% & 0.08 & 0.43 & 0.62 & 1.00 & 0.18 & 1.07 & 1.79 & 2.90 & 0.28 & 2.04 & 2.90 & 3.62 \\
SMT & 2 & 0\% & 0.17 & 0.56 & 0.90 & 1.36 & 0.23 & 1.13 & 1.78 & 2.96 & 0.38 & 1.60 & 2.28 & 3.52 \\
\midrule
AMT-M & 0 & \cellcolor{green!15}94\% & 0.15 & \cellcolor{green!15}0.45 & \cellcolor{green!15}0.59 & \cellcolor{green!15}0.76 & 0.23 & 0.84 & 1.27 & 1.85 & 0.49 & 1.43 & 1.89 & 2.73 \\
AMT-M & 1 & \cellcolor{green!40}100\% & 0.17 & 0.57 & 0.75 & \cellcolor{green!15}1.01 & 0.21 & 0.75 & 1.10 & 1.59 & 0.40 & 1.14 & 1.64 & \cellcolor{green!15}2.20 \\
AMT-M & 2 & \cellcolor{green!15}94\% & 0.15 & 0.49 & 0.67 & \cellcolor{green!15}0.96 & 0.19 & 0.92 & 1.35 & 1.99 & 0.52 & 1.41 & 2.08 & 2.90 \\
\midrule
AMT-B & 0 & \cellcolor{green!15}96\% & 0.08 & \cellcolor{green!15}0.32 & \cellcolor{green!15}0.43 & \cellcolor{green!15}0.59 & 0.13 & 0.46 & 0.78 & 1.39 & 0.33 & 0.85 & \cellcolor{green!15}1.04 & \cellcolor{green!15}1.41 \\
AMT-B & 1 & \cellcolor{green!15}96\% & 0.11 & \cellcolor{green!15}0.38 & \cellcolor{green!15}0.51 & \cellcolor{green!15}0.74 & 0.17 & 0.72 & 1.10 & 1.69 & 0.30 & 1.17 & 1.71 & \cellcolor{green!15}2.27 \\
AMT-B & 2 & \cellcolor{green!15}99\% & 0.10 & \cellcolor{green!15}0.37 & \cellcolor{green!15}0.50 & \cellcolor{green!15}0.66 & 0.20 & 0.62 & 0.91 & 1.41 & 0.36 & 1.19 & 1.51 & \cellcolor{green!15}1.95 \\
\midrule
& & & \multicolumn{4}{c}{$|\Delta \delta_{e,max}|$ [deg]} &\multicolumn{4}{c}{$|\Delta \delta_{a,max}|$ [deg]} & \multicolumn{4}{c}{$|\tau_z|$ [N]}\\
\midrule
Baseline   & - & 79\% & \cellcolor{green!40}0.00 & \cellcolor{green!40}0.01 & \cellcolor{green!40}0.01 & \cellcolor{green!40}0.06 & \cellcolor{green!40}0.00 & \cellcolor{green!40}0.01 & \cellcolor{green!40}0.03 & \cellcolor{green!40}0.06 & \cellcolor{green!40}0.31 & \cellcolor{green!40}16.11 & 71.70 & 98.1 \\
\midrule
DR (Add.) & 0 & \cellcolor{green!15}93\% & 0.00 & 0.10 & 0.21 & 0.37 & 0.00 & 0.07 & 0.12 & 0.21 & 5.49 & 18.25 & \cellcolor{green!15}53.15 & \cellcolor{green!15}89.0 \\
DR (Add.) & 1 & \cellcolor{green!15}93\% & 0.02 & 0.26 & 0.43 & 0.69 & 0.01 & 0.13 & 0.22 & 0.38 & 8.46 & 25.73 & \cellcolor{green!15}50.76 & \cellcolor{green!15}88.2 \\
DR (Add.) & 2 & \cellcolor{green!15}94\% & 0.00 & 0.21 & 0.38 & 0.66 & 0.00 & 0.15 & 0.26 & 0.43 & 6.24 & 31.23 & \cellcolor{green!15}60.88 & \cellcolor{green!15}93.0 \\
\midrule
DR & 0 & \cellcolor{green!40}100\% & 0.02 & 0.43 & 0.56 & 0.71 & 0.01 & 0.27 & 0.42 & 0.63 & 6.15 & 23.85 & \cellcolor{green!40}33.44 & \cellcolor{green!40}51.1 \\
DR & 1 & \cellcolor{green!15}99\% & 0.01 & 0.39 & 0.60 & 0.84 & 0.00 & 0.21 & 0.34 & 0.54 & 6.28 & 30.23 & \cellcolor{green!15}49.73 & \cellcolor{green!15}72.8 \\
DR & 2 & \cellcolor{green!15}98\% & 0.02 & 0.48 & 0.66 & 0.86 & 0.01 & 0.26 & 0.46 & 0.88 & 4.95 & 23.49 & \cellcolor{green!15}39.33 & \cellcolor{green!15}67.0 \\
\midrule
RR & 0 & \cellcolor{green!15}98\% & 0.09 & 0.56 & 0.68 & 0.83 & 0.04 & 0.31 & 0.46 & 0.65 & 10.95 & 37.90 & \cellcolor{green!15}57.59 & 100.6 \\
RR & 1 & \cellcolor{green!15}96\% & 0.13 & 0.66 & 0.79 & 0.92 & 0.09 & 0.50 & 0.64 & 0.78 & 15.56 & 56.70 & 75.46 & 107.0 \\
RR & 2 & \cellcolor{green!15}95\% & 0.06 & 0.55 & 0.75 & 0.95 & 0.04 & 0.36 & 0.48 & 0.63 & 10.26 & 37.39 & \cellcolor{green!15}55.20 & \cellcolor{green!15}85.7 \\
\midrule
SMT & 0 & 0\% & 0.04 & 0.85 & 1.01 & 1.06 & 0.02 & 0.32 & 0.54 & 0.93 & 3.31 & 53.46 & 76.66 & 126.8 \\
SMT & 1 & 0\% & 0.08 & 0.70 & 0.84 & 0.96 & 0.03 & 0.47 & 0.71 & 1.06 & 3.23 & 47.38 & 78.06 & 127.8 \\
SMT & 2 & 0\% & 0.16 & 0.80 & 0.90 & 0.98 & 0.14 & 1.06 & 1.07 & 1.07 & 17.23 & 69.22 & 102.72 & 152.5 \\
\midrule
AMT-M & 0 & \cellcolor{green!15}94\% & 0.03 & 0.35 & 0.55 & 0.82 & 0.03 & 0.37 & 0.53 & 0.72 & 15.22 & 47.83 & \cellcolor{green!15}71.20 & 115.6 \\
AMT-M & 1 & \cellcolor{green!40}100\% & 0.05 & 0.47 & 0.60 & 0.76 & 0.04 & 0.33 & 0.45 & 0.61 & 13.24 & 54.50 & 74.81 & 103.8 \\
AMT-M & 2 & \cellcolor{green!15}94\% & 0.21 & 0.77 & 0.91 & 1.03 & 0.14 & 0.67 & 0.91 & 1.06 & 14.74 & 64.18 & 90.91 & 139.2 \\
\midrule
AMT-B & 0 & \cellcolor{green!15}96\% & 0.01 & 0.30 & 0.43 & 0.58 & 0.01 & 0.25 & 0.40 & 0.62 & 7.69 & 30.31 & \cellcolor{green!15}49.38 & \cellcolor{green!15}89.6 \\
AMT-B & 1 & \cellcolor{green!15}96\% & 0.12 & 0.65 & 0.79 & 0.95 & 0.08 & 0.42 & 0.56 & 0.77 & 10.37 & 47.69 & \cellcolor{green!15}71.30 & 107.5 \\
AMT-B & 2 & \cellcolor{green!15}99\% & 0.11 & 0.62 & 0.78 & 0.96 & 0.05 & 0.34 & 0.50 & 0.72 & 16.36 & 51.86 & \cellcolor{green!15}69.21 & 100.8 \\
\bottomrule
\end{longtable}
}
\end{table}

\section{Individual Learning Curves Under Nominal Conditions}
\label{app:individual_learning_curves}

Plots throughout the paper show the IQM for the returns, which remove outliers, e.g., failed runs. Here we plot individual learning curves for training under nominal conditions without outlier filtering (experiments from Section~\ref{sec:rl_alg_selection}). For each training run, we average the performance over ten evaluation episodes. For comparison, we plot the performance of the baseline controller, no control, and the IQM of all training runs. Individual learning curves for MR.Q indicate that the training process is slightly unstable, but all runs surpass the baseline controller at some point.

\begin{figure}
\centering
\includegraphics[width=\textwidth]{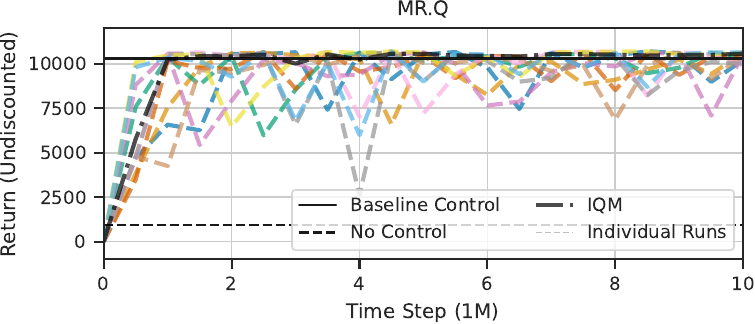}
\caption{Individual learning curves of MR.Q (Only RL).}
\label{fig:learning_curves_individual_mrq}
\end{figure}

\begin{figure}
\centering
\includegraphics[width=\textwidth]{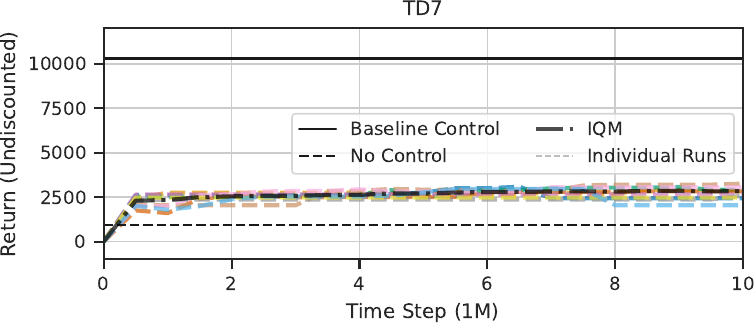}
\caption{Individual learning curves of TD7 (Only RL).}
\label{fig:learning_curves_individual_td7}
\end{figure}

\begin{figure}
\centering
\includegraphics[width=\textwidth]{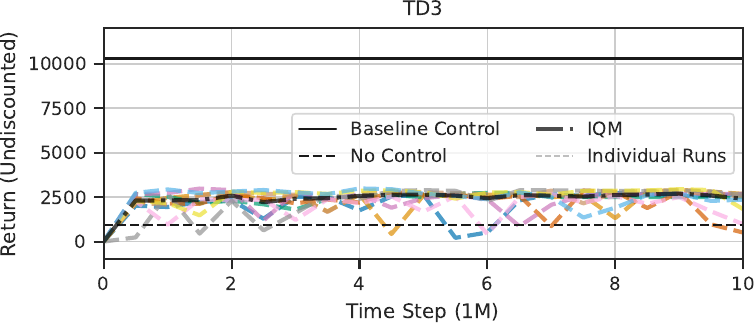}
\caption{Individual learning curves of TD3 (Only RL).}
\label{fig:learning_curves_individual_td3}
\end{figure}

\begin{figure}
\centering
\includegraphics[width=\textwidth]{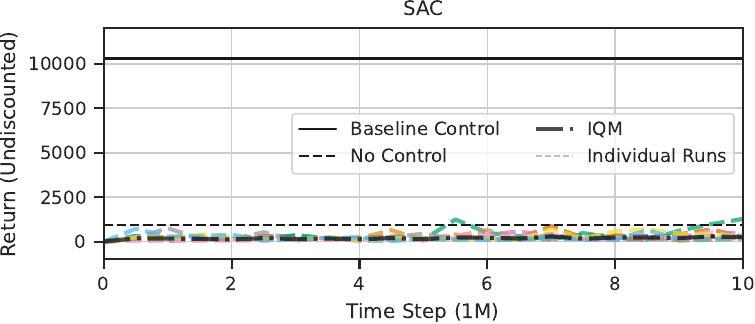}
\caption{Individual learning curves of SAC (Only RL).}
\label{fig:learning_curves_individual_sac}
\end{figure}

\begin{figure}
\centering
\includegraphics[width=\textwidth]{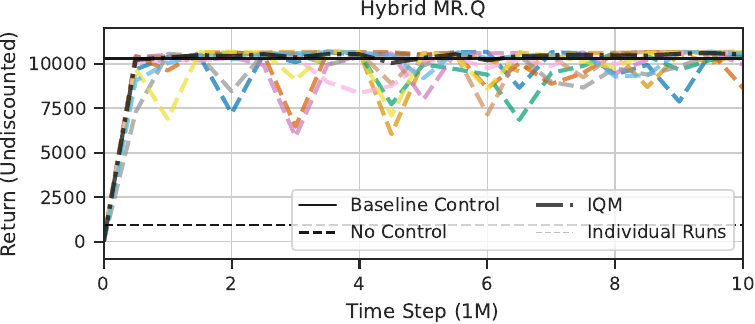}
\caption{Individual learning curves of MR.Q (Additive Hybrid).}
\label{fig:learning_curves_individual_hybrid_mrq}
\end{figure}

\begin{figure}
\centering
\includegraphics[width=\textwidth]{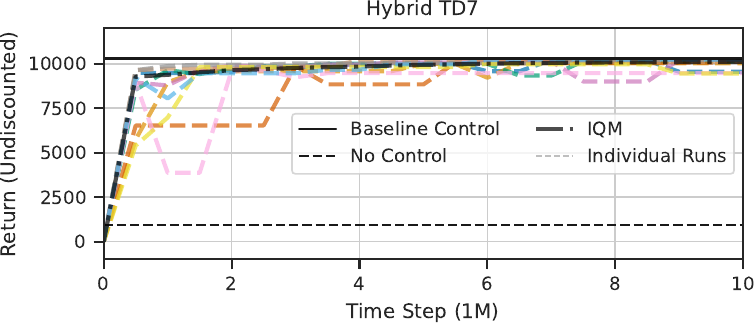}
\caption{Individual learning curves of TD7 (Additive Hybrid).}
\label{fig:learning_curves_individual_hybrid_td7}
\end{figure}

\begin{figure}
\centering
\includegraphics[width=\textwidth]{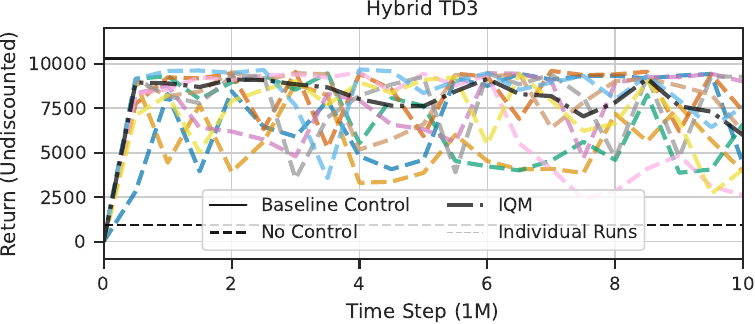}
\caption{Individual learning curves of TD3 (Additive Hybrid).}
\label{fig:learning_curves_individual_hybrid_td3}
\end{figure}

\begin{figure}
\centering
\includegraphics[width=\textwidth]{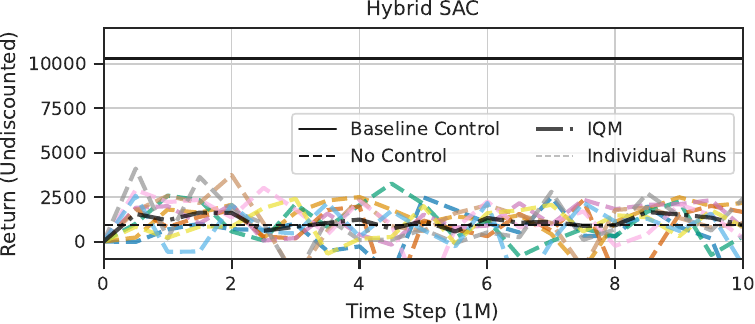}
\caption{Individual learning curves of SAC (Additive Hybrid).}
\label{fig:learning_curves_individual_hybrid_sac}
\end{figure}

\begin{figure}
\centering
\includegraphics[width=\textwidth]{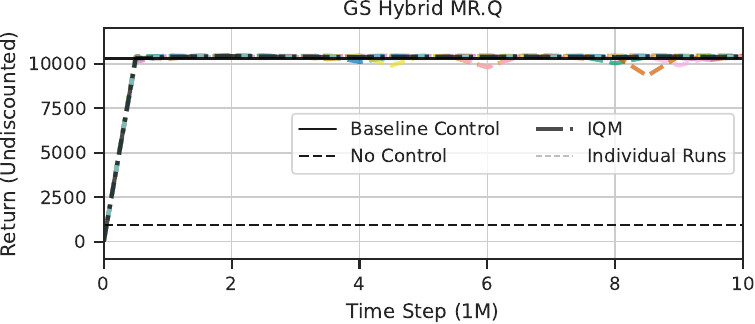}
\caption{Individual learning curves of MR.Q (Gain-Scheduling Hybrid).}
\label{fig:learning_curves_individual_gs_hybrid_mrq}
\end{figure}

\end{document}